\documentclass[sigconf]{acmart}

\usepackage{microtype}
\usepackage{graphicx}
\usepackage{caption}
\usepackage{subcaption}
\usepackage{float}
\usepackage{epsfig}
\usepackage{stfloats}
\usepackage{capt-of}
\usepackage{multirow}
\usepackage{booktabs}
\usepackage{amsmath}

\usepackage{amssymb}
\usepackage{mathtools}
\usepackage{amsthm}
\usepackage{bm}
\usepackage{algorithm}
\usepackage{algpseudocode}

\AtBeginDocument{%
  }

\renewcommand\footnotetextcopyrightpermission[1]{}

\acmConference[Preprint]{arXiv}{2026}{}

\begin{document}

\title{A Controlled Diagnostic Study of Hardware-Induced Distortions in Hardware-Aware Training}

\author{Yunxuan Fang}
\affiliation{
  \institution{Beihang University}
  \city{Beijing}
  \country{China}
}
\email{micro@buaa.edu.cn}

\author{Xinhe Wang}
\affiliation{
  \institution{Beihang University}
  \city{Beijing}
  \country{China}
}
\email{xinhe@buaa.edu.cn}


\begin{abstract}
Hardware-aware training (HAT) is widely used to improve the robustness of neural networks on non-ideal AI accelerators, such as analog in-memory computing (IMC) systems. However, not all hardware-induced distortions are equally compensable by training. This paper presents a diagnostic framework that models hardware non-idealities as structured perturbations of the forward operator and evaluates their compatibility with gradient-based optimization. We analyze six representative perturbation classes—read noise, variability, drift, stuck-at faults, IR-drop, and ADC discretization—and identify three key diagnostics: gradient expectation consistency, bounded gradient variance, and non-degenerate sensitivity. Our results show a clear separation between perturbations that can be compensated by HAT and those that consistently break optimization. This provides practical guidance for hardware-software co-design, clarifying which non-idealities can be addressed at the training level and which require circuit-, architecture-, or calibration-level mitigation.
This study should be interpreted as a controlled empirical analysis under vanilla forward-perturbation HAT, rather than as a universal theory of hardware-aware training.
\end{abstract}



\maketitle

\section{Introduction}

Emerging AI accelerators such as analog in-memory computing (IMC) systems offer major gains in energy efficiency and throughput, but their benefits are often limited by hardware non-idealities including device variability, conductance drift, stuck-at faults, IR-drop, and finite-precision readout. These effects distort the effective forward operator seen by a neural network and can significantly degrade accuracy if the model is trained only under ideal software assumptions.

Hardware-aware training (HAT) is a widely used strategy for addressing this problem. By injecting hardware distortions during training, HAT allows model parameters to adapt to non-ideal execution. In practice, however, HAT is not uniformly effective: some non-idealities can be compensated successfully, while others consistently destabilize optimization. This motivates a practical hardware--software co-design question: in a given HAT setup, which distortions appear compatible with training-time compensation, and which may require circuit-, architecture-, or calibration-level mitigation?

Existing studies typically examine individual non-idealities in isolation and evaluate whether a particular compensation method improves final accuracy. As a result, the literature offers many perturbation-specific observations but less understanding of why certain hardware effects remain learnable under training while others do not. Moreover, higher physical fidelity does not necessarily imply better training behavior: a more realistic hardware model may introduce stronger couplings or non-smooth effects that make gradient-based optimization less stable.

In this work, we study this question through a structured abstraction of hardware-induced perturbations. Rather than focusing on a single device mechanism, we organize diverse non-idealities according to how they interact with the forward operator and examine their compatibility with gradient-based training. Across six representative perturbation classes spanning read noise, variability, drift, stuck-at faults, IR-drop, and ADC discretization, we observe that compensability is closely associated with several recurring optimization diagnostics, including gradient expectation consistency, bounded gradient variance, and non-degenerate sensitivity.

These observations lead to a practical perspective on HAT. Perturbations with stable gradient statistics are often amenable to training-time compensation, whereas strongly coupled or non-smooth perturbations tend to break optimization and therefore require mitigation beyond the training loop. Our results provide a diagnostic view for interpreting when vanilla HAT succeeds or fails in the studied perturbation regimes, and suggest when hardware-side intervention may be preferable.

Our main contributions are summarized as follows:
\begin{itemize}
    \item We propose a structured abstraction of hardware non-idealities that captures common interaction patterns between hardware-induced perturbations and neural network forward operators.
    \item We identify recurring optimization diagnostics that empirically distinguish learnable perturbations from those that lead to training instability.
    \item We show that these diagnostics help separate perturbations that are compatible with vanilla training-time compensation from those that exhibit optimization collapse in our controlled experiments.
    \item We translate these observations into hardware--software co-design guidance, clarifying which non-idealities are suitable for training-time compensation and which instead call for circuit-, architecture-, or calibration-level mitigation.
\end{itemize}

\section{Related Work}

\subsection{Hardware-Aware Training}

Hardware-aware training (HAT) has emerged as a widely adopted strategy for bridging the gap between idealized software models and non-ideal hardware implementations~\cite{rasch2023hardware, zhang2020neuro}. Early work in this direction focused on quantization-aware training (QAT), where weight and activation discretization are incorporated during training to improve robustness to fixed-point arithmetic~\cite{hubara2018quantized, Jacob2018Quantization, chen2025efficientqat}. Similar ideas have been extended to a broader range of hardware platforms, including analog and mixed-signal accelerators, where device variability, conductance drift, read noise, and circuit non-idealities must be accounted for during training~\cite{lanza2025growing, aguirre2024hardware}.

For analog in-memory computing (IMC) systems, HAT is often implemented by injecting simulated hardware perturbations into the forward pass during training~\cite{sebastian2020memory, rasch2021flexible}. This allows networks to adapt their parameters to compensate for device-level variations and circuit-induced distortions. Related techniques also include calibration-based approaches and joint optimization of network parameters with hardware parameters~\cite{lin2019performance, zhang2020fast}.

Despite their empirical effectiveness, most existing HAT approaches are designed for specific hardware effects and rely on heuristic perturbation models. As a result, they provide limited insight into the broader question of when training can successfully compensate for hardware-induced distortions and when such compensation becomes fundamentally difficult.

\vspace{-0.5\baselineskip}

\subsection{Modeling and Learnability of Hardware Perturbations}

A large body of work has focused on modeling non-idealities in emerging computing substrates such as memristive crossbar arrays, SRAM-based in-memory accelerators, and neuromorphic hardware~\cite{sebastian2020memory, rasch2021flexible, davies2018loihi}. These studies often develop detailed physical models capturing effects such as device mismatch, retention drift, IR-drop, and sensing noise. While such models improve simulation fidelity, they also introduce complex couplings between weights, activations, and circuit states, which can significantly complicate gradient-based training.

To reduce modeling complexity, several works have proposed simplified abstractions of hardware perturbations, such as additive noise models, multiplicative variability models, or column-wise attenuation approximations for IR-drop~\cite{joshi2020accurate, rasch2023hardware, lanza2025growing}. These abstractions make hardware-aware training more tractable but are typically motivated by simulation convenience rather than by an explicit analysis of learnability.

From the optimization perspective, classical results on stochastic gradient descent show that convergence can be guaranteed under unbiased gradient estimates and bounded variance conditions~\cite{bottou2012stochastic, zinkevich2010parallelized}. More recent studies have examined the effect of structured gradient corruption and noise on training dynamics~\cite{simsekli2020fractional}. However, these analyses generally assume perturbation structures that are already known to be learnable and do not provide a systematic diagnostic framework for distinguishing compensable and non-compensable perturbations in hardware-aware training.

In contrast, our work adopts a diagnostic perspective. Instead of focusing on improving perturbation modeling fidelity or designing perturbation-specific mitigation techniques, we investigate how different perturbation structures influence the optimization dynamics of gradient-based training. By connecting perturbation structure with observable gradient statistics during training, our framework provides a practical lens for interpreting heterogeneous outcomes of hardware-aware training across diverse perturbation classes.

\section{Hardware Non-Ideality Abstraction for HAT Analysis}
\label{sec:perturbation-framework}

\subsection{Learning under Structured Operator Perturbations}

Analog and mixed-signal AI accelerators exhibit diverse hardware non-idealities, including read noise, drift, stuck-at faults, IR-drop, and discretization. Although these effects differ in physical origin, they all modify the effective forward operator seen during inference and hardware-aware training.

In this work, we study six representative perturbation classes chosen to capture recurring algebraic interaction patterns between perturbations and the forward operator, as well as structures commonly encountered in analog and mixed-signal hardware. Rather than pursuing a universal physical model, we adopt an abstraction-first view: the goal is to organize perturbations by how they interact with trainable parameters and inputs, so that their learnability under hardware-aware training (HAT) can later be analyzed in a common framework.

\subsection{Unified Forward Operator}

Consider an ideal linear layer (including convolution) written as $\mathbf{z}=\mathbf{W}\mathbf{x}$, $\mathbf{y}=\phi(\mathbf{z})$, where $\mathbf{W}$ denotes trainable parameters, $\mathbf{x}$ is the input activation, and $\phi(\cdot)$ denotes subsequent nonlinearities.

We model hardware-perturbed computation as $\tilde{\mathbf{z}}=\mathcal{G}(\mathbf{W},\mathbf{x};\xi,t)$, $\tilde{\mathbf{y}}=\phi(\tilde{\mathbf{z}})$, where $\xi$ denotes stochastic perturbation variables and $t$ denotes deterministic time- or cycle-dependent effects. For analysis, we express perturbations through an effective operator in the weight domain: $\tilde{\mathbf{z}}=\mathbf{W}_{\mathrm{eff}}(\mathbf{W},\mathbf{x};\xi,t)\,\mathbf{x}$.

When a perturbation naturally acts on activations or outputs, we reinterpret it as an equivalent transformation in $\mathbf{W}_{\mathrm{eff}}$ so that diverse perturbations can be analyzed within one common notation.

\subsection{Hardware-Aware Training under Structured Perturbations}

Under this abstraction, HAT is implemented by sampling perturbations during the forward pass while optimizing the underlying software weights. Concretely, each training step constructs an effective perturbed operator $\mathbf{W}_{\mathrm{eff}}(\mathbf{W},\mathbf{x};\xi,t)$, performs forward computation with the perturbed model, and updates the trainable parameters $\mathbf{W}$ through backpropagation.

The training objective is
\[
\mathcal{L}(\mathbf{W})
=
\mathbb{E}_{(x,y),\xi,t}
\big[\ell(f(x;\mathbf{W},\xi,t),y)\big]
+
\lambda_{\mathrm{reg}}\mathcal{L}_{\mathrm{reg}}(\mathbf{W}),
\]
where $\xi$ and $t$ denote stochastic and time-dependent hardware effects, respectively. In all experiments, perturbations are injected only in the forward computation, while gradients are taken with respect to the trainable software weights.

A perturbation class is regarded as compensable if offline gradient-based optimization can adapt $\mathbf{W}$ so that the trained model mitigates the corresponding inference-time distortion.

\subsection{Six Algebraic Perturbation Classes}

We categorize perturbations by their algebraic interaction pattern with the forward operator. The taxonomy is intentionally compact: its role is to define the perturbation space to be diagnosed in Section~\ref{sec:learnability-boundary}, rather than to fully analyze learnability here. The six classes are not intended to exhaust all hardware effects; rather, they capture recurrent operator-level structures that appear across several practically important non-idealities.

\textbf{Additive: read noise / sense amplifier noise.}
\[
\mathbf{W}_{\mathrm{eff}}=\mathbf{W}+\mathbf{E}_A,
\qquad
\mathbb{E}[\mathbf{E}_A]=\mathbf{0}.
\]

\textbf{Multiplicative perturbations: programming variability / retention drift.}
\[
\mathbf{W}_{\mathrm{eff}}=\mathbf{A}(\xi,t)\odot \mathbf{W}.
\]

\textbf{Projection: stuck-at / dead-cell / write-failure induced freezing.}
\[
\mathbf{W}_{\mathrm{eff}}
=
\mathbf{S}\odot \mathbf{W}
+
(\mathbf{1}-\mathbf{S})\odot \mathbf{C}.
\]

\textbf{Input-dependent structured scaling: simplified IR-drop for large crossbar arrays.}
\[
\tilde{\mathbf{z}}
=
\mathbf{W}\big(\mathbf{D}(\mathbf{W},\mathbf{x};\xi)\mathbf{x}\big)
\quad \Leftrightarrow \quad
\mathbf{W}_{\mathrm{eff}}=\mathbf{W}\mathbf{D}(\mathbf{W},\mathbf{x};\xi).
\]

\textbf{Strongly coupled nonlinear: high-fidelity IR-drop / parasitic coupling models.}
\[
\tilde{\mathbf{z}}
=
\mathbf{W}\mathbf{x}
+
\Delta(\mathbf{W},\mathbf{x};\xi).
\]

\textbf{Discretization: ADC / low-precision readout path.}
\[
\tilde{\mathbf{z}}=Q(\mathbf{z}),
\qquad
\partial Q(z)/\partial z = 0 \ \text{for a.e. } z.
\]

\subsection{Evaluation Protocol}

All experiments are conducted on ResNet-20 with CIFAR-10 under a baseline hardware configuration that already includes multiple non-idealities. Without compensation, this baseline hardware yields 35.77\% accuracy. When evaluating a given perturbation class, we vary only its corresponding perturbation parameter while holding all other non-idealities fixed at their baseline values. This isolates the effect of perturbation structure on learnability. 
Our empirical study is intentionally controlled and focuses on one representative vision architecture and dataset, with the goal of isolating the role of perturbation structure rather than claiming architecture-universal thresholds.
We therefore treat the results as evidence for perturbation-structure effects under a fixed training protocol, not as a claim that the same quantitative boundaries transfer unchanged across architectures or tasks.
This setup is intended to mimic a practically non-ideal operating point rather than an isolated single-noise simulation, so that the measured compensation behavior reflects perturbation structure under realistic hardware degradation.

\section{Diagnostic Framework and Empirical Validation}
\label{sec:learnability-boundary}

We now turn to the central question of this work: when can gradient-based optimization compensate for hardware-induced perturbations injected into the forward operator? 

\subsection{Diagnostic Patterns for Compensation}

Consider the training objective defined over perturbed forward passes:
\[
\mathcal{L}(\mathbf{W})=
\mathbb{E}_{(\mathbf{x},y),\xi,t}
\,\ell\!\left(f_{\mathcal{G}}(\mathbf{x};\mathbf{W},\xi,t),y\right).
\]
We repeatedly observe that perturbations remain compensable in our experiments when the induced optimization dynamics preserve the following three diagnostic proxies.

\textbf{Gradient Expectation Consistency.}
The expected perturbed gradient remains aligned with the gradient of a stable surrogate objective:
\[
\mathbb{E}_{\xi,t}
\left[
\frac{\partial \tilde{\mathbf{z}}}{\partial \mathbf{W}}
\right]
\approx
\frac{\partial \mathbb{E}_{\xi,t}[\tilde{\mathbf{z}}]}{\partial \mathbf{W}}.
\]
Intuitively, the perturbation should preserve a sufficiently stable algebraic relationship between $\mathbf{W}$ and $\mathbf{x}$.

\textbf{Bounded Gradient Variance.}
The perturbation-induced gradient noise remains controlled:
\[
\mathrm{Var}_{\xi,t}
\left(
\frac{\partial \tilde{\mathbf{z}}}{\partial \mathbf{W}}
\right)
<\infty.
\]
When the effective variance becomes excessively large in practice, SGD updates become unstable and compensation fails.

\textbf{Non-degenerate Sensitivity.}
The perturbed forward operator retains non-trivial sensitivity to trainable parameters:
\[
\left\|
\frac{\partial \tilde{\mathbf{z}}}{\partial \mathbf{W}}
\right\|
\not\equiv 0.
\]
If this sensitivity collapses, optimization loses access to task-relevant gradient signals even if the forward pass remains well-defined.

They are not proposed as formal convergence guarantees; instead, they summarize measurable symptoms observed in accuracy and gradient-dynamics curves. They should therefore be interpreted as diagnostic criteria for characterizing observed optimization behavior in HAT, rather than as a complete convergence theory for all hardware perturbations. Their role is to provide a compact lens for explaining why some perturbation structures remain learnable under HAT while others do not.

\subsection{Three Compensation Regimes}

\begin{figure*}[htb]
\centering
\begin{subfigure}[b]{0.32\textwidth}
\centering
\includegraphics[width=\linewidth]{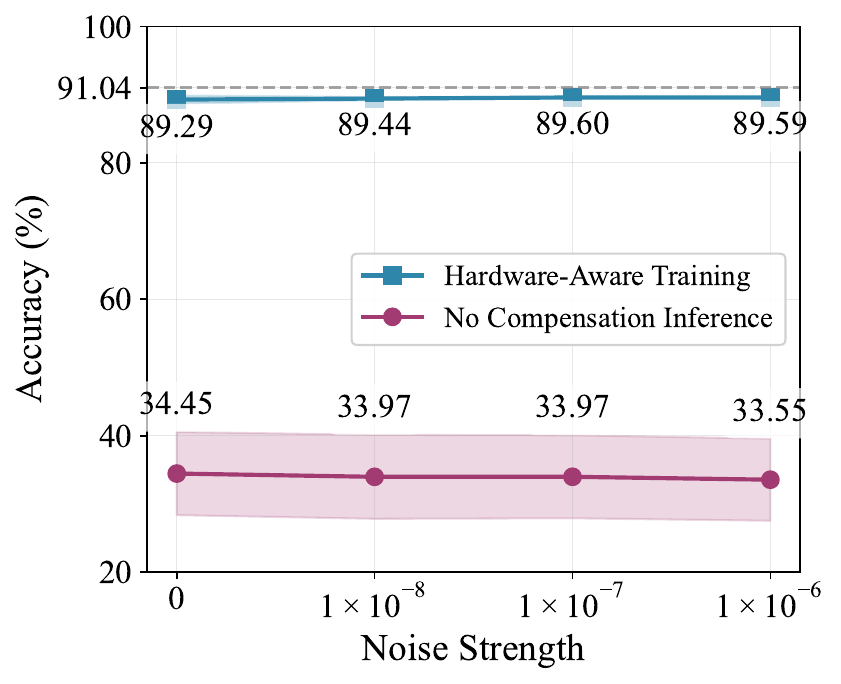}
\caption{Additive perturbations}
\label{fig:additive}
\end{subfigure}
\hfill
\begin{subfigure}[b]{0.32\textwidth}
\centering
\includegraphics[width=\linewidth]{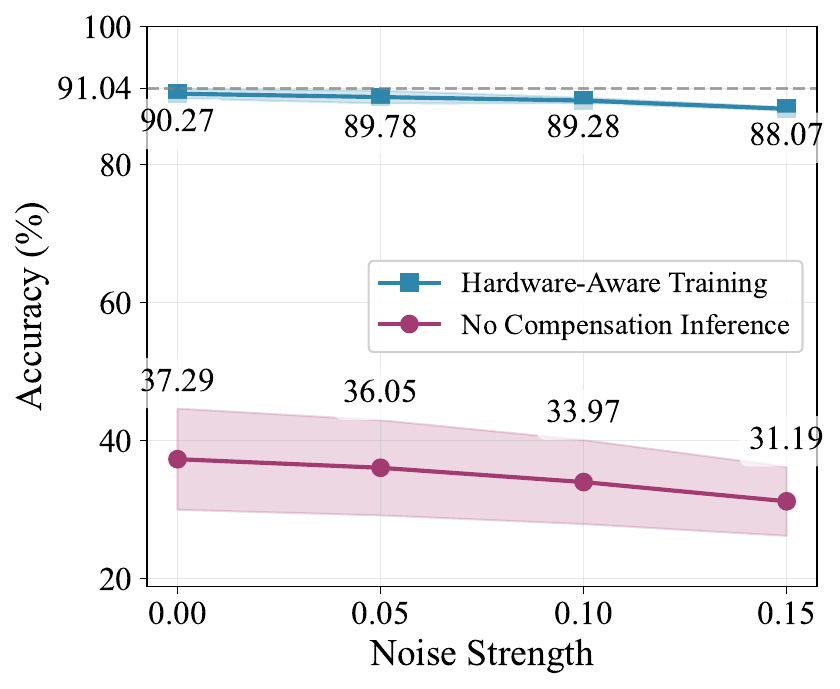}
\caption{Multiplicative perturbations}
\label{fig:multiplicative}
\end{subfigure}
\hfill
\begin{subfigure}[b]{0.32\textwidth}
\centering
\includegraphics[width=\linewidth]{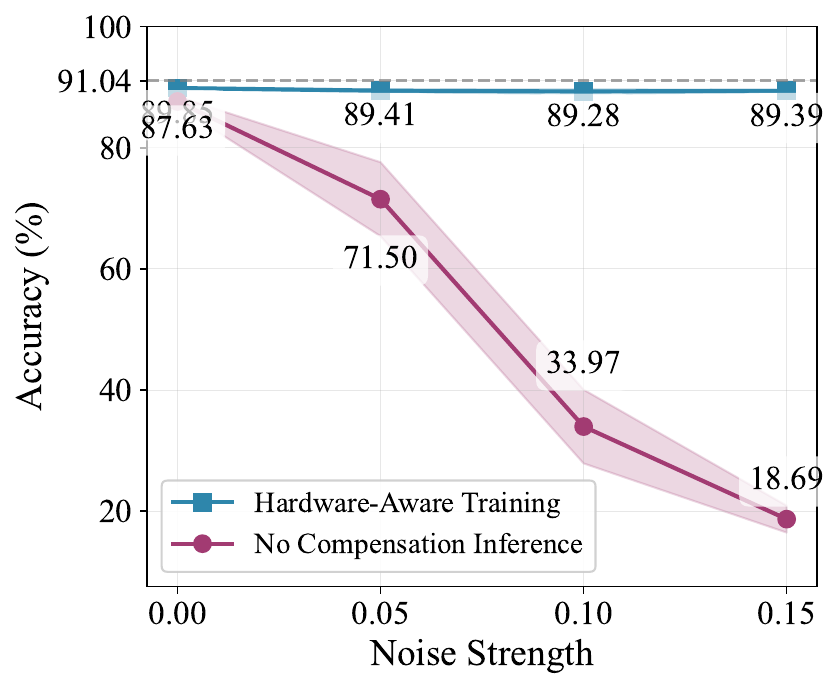}
\caption{Projection perturbations}
\label{fig:projection}
\end{subfigure}
\hfill
\begin{subfigure}[b]{0.32\textwidth}
\centering
\includegraphics[width=\linewidth]{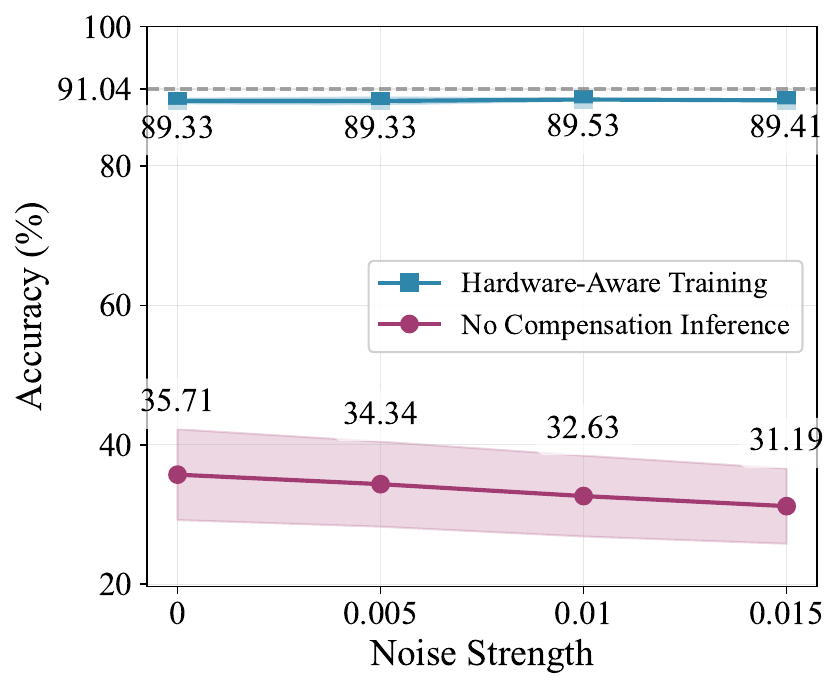}
\caption{Input-dependent scaling}
\label{fig:input_scaling}
\end{subfigure}
\hfill
\begin{subfigure}{0.32\textwidth}
\centering
\includegraphics[width=\linewidth]{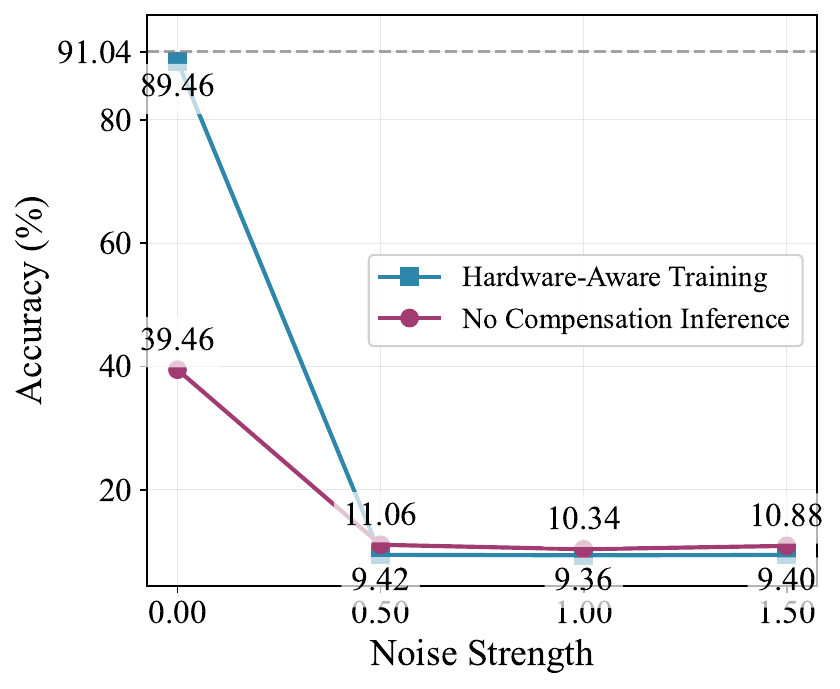}
\caption{Strongly Coupled Nonlinear}
\label{fig:IR-drop}
\end{subfigure}
\hfill
\begin{subfigure}{0.32\textwidth}
\centering
\includegraphics[width=\linewidth]{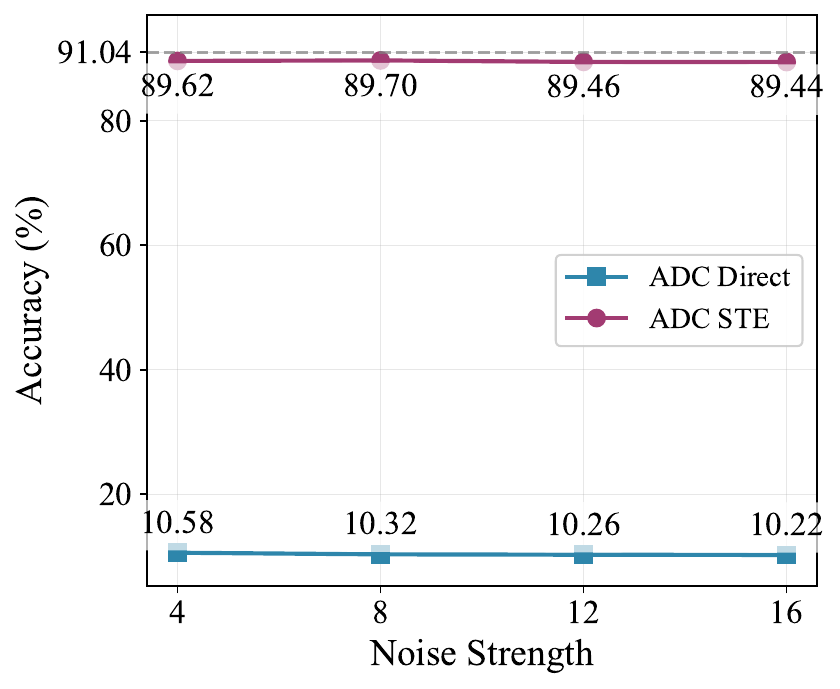}
\caption{Discretization Operators}
\label{fig:ADC}
\end{subfigure}
\caption{\textbf{Accuracy under six perturbation classes.} (a) Additive read noise $\sigma_r$. (b) Multiplicative variability $\sigma_v$. (c) Projection perturbations with stuck-at ratio $\rho$. (d) Input-dependent structured scaling with IR-drop strength $\beta$. (e) Strongly coupled nonlinear IR-drop. (f) ADC discretization. HAT recovers accuracy for the first four classes but fails for strongly coupled nonlinear perturbations and direct discretization. Experiments use ResNet-20 on CIFAR-10.}
\label{fig:main_figure}
\end{figure*}

\begin{figure*}[htb]
\centering
\begin{subfigure}[b]{0.32\textwidth}
\centering
\includegraphics[width=\linewidth]{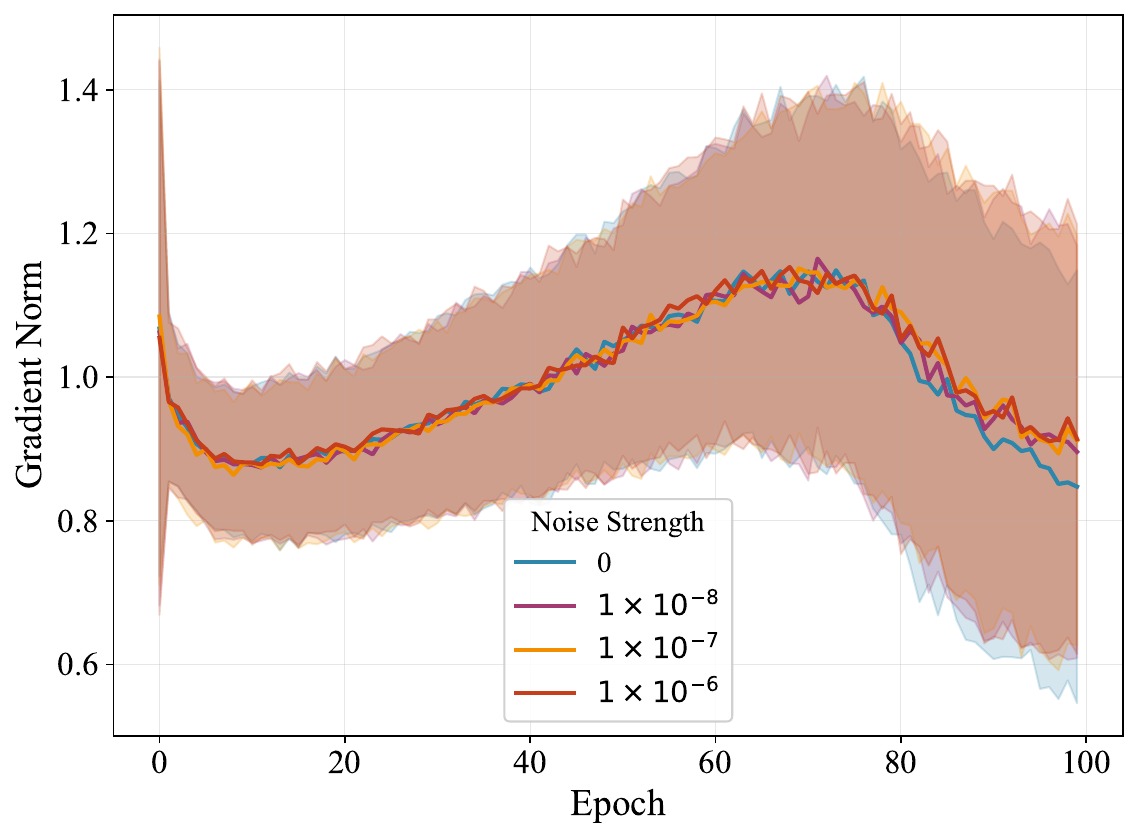}
\caption{Additive: $\sigma_r$}
\label{fig:grad-additive}
\end{subfigure}
\hfill
\begin{subfigure}[b]{0.32\textwidth}
\centering
\includegraphics[width=\linewidth]{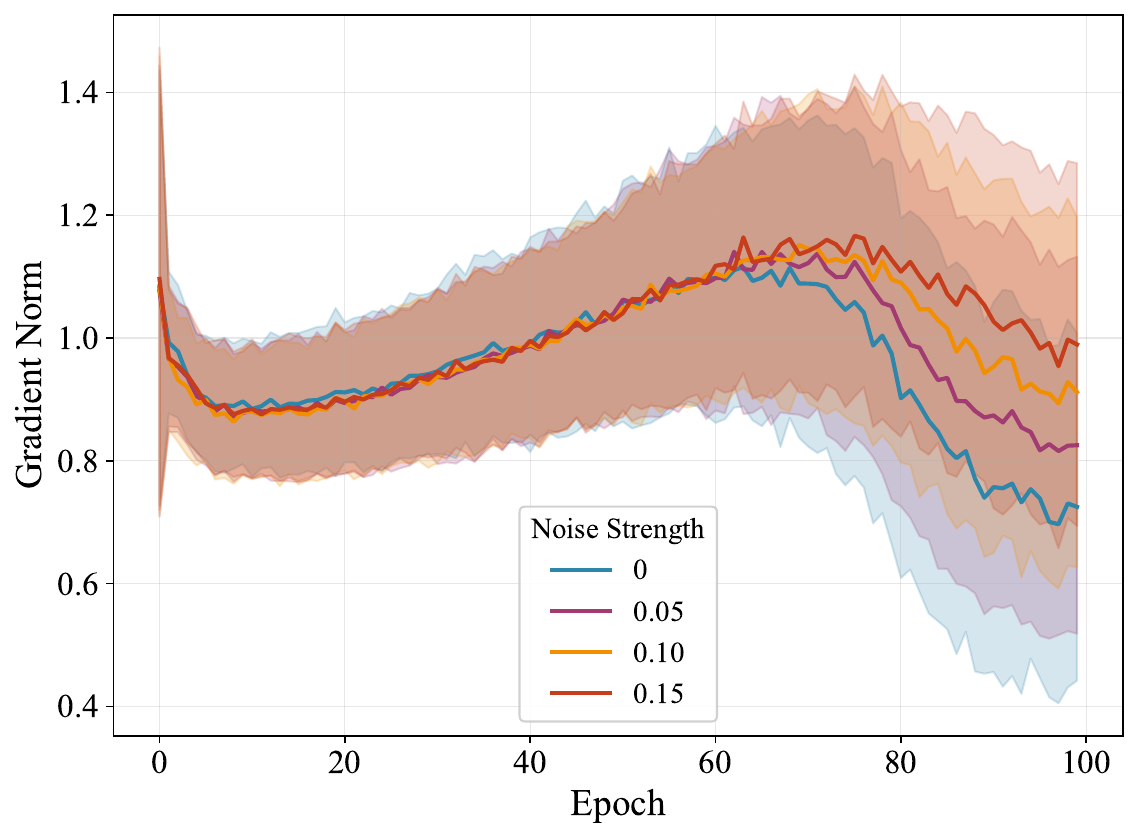}
\caption{Multiplicative: $\sigma_v$}
\label{fig:grad-multiplicative}
\end{subfigure}
\hfill
\begin{subfigure}[b]{0.32\textwidth}
\centering
\includegraphics[width=\linewidth]{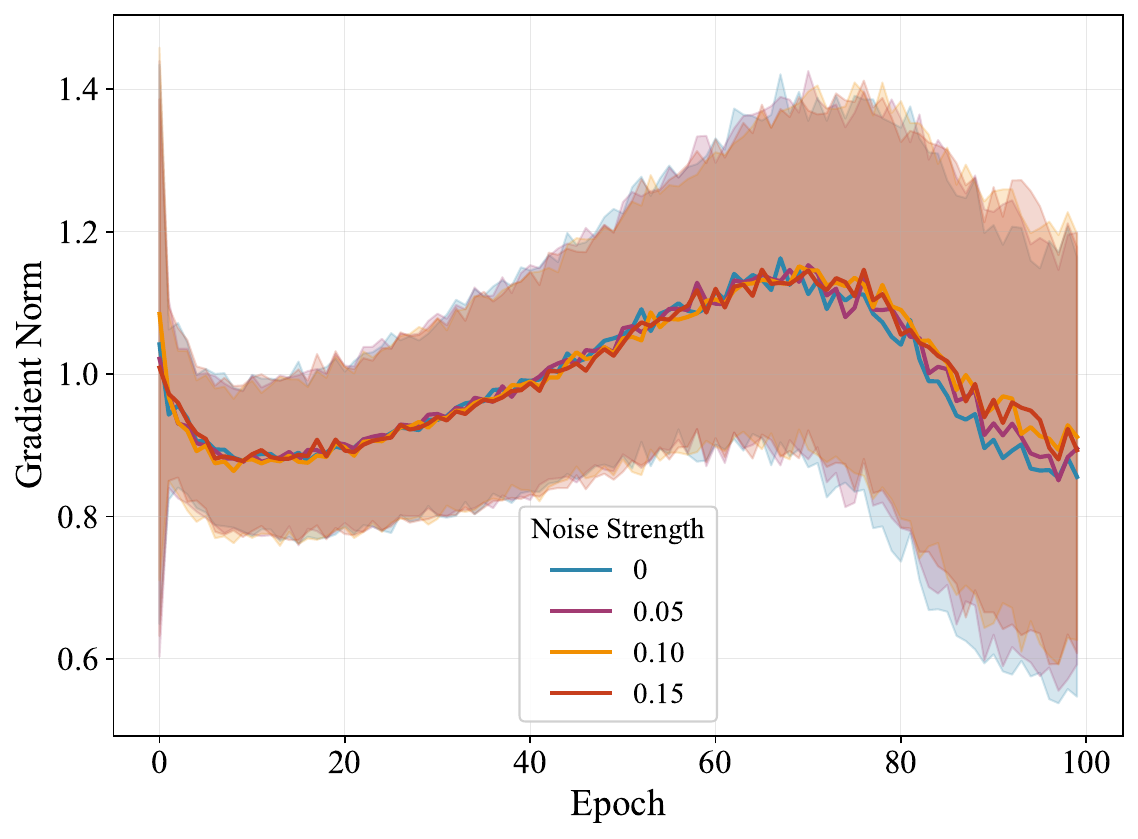}
\caption{Projection: $\rho$}
\label{fig:grad-projection}
\end{subfigure}
\hfill
\begin{subfigure}[b]{0.32\textwidth}
\centering
\includegraphics[width=\linewidth]{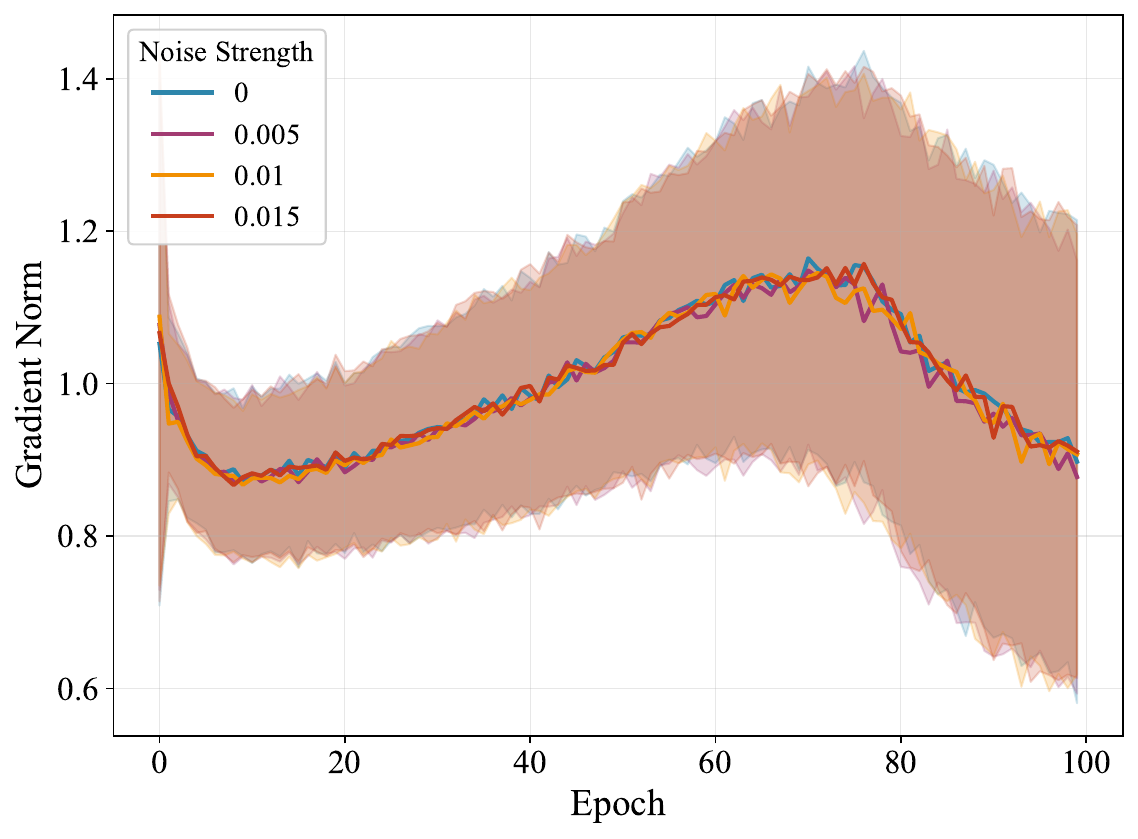}
\caption{Input scaling: $\beta$}
\label{fig:grad-input-scaling}
\end{subfigure}
\hfill
\begin{subfigure}{0.32\textwidth}
\centering
\includegraphics[width=\linewidth]{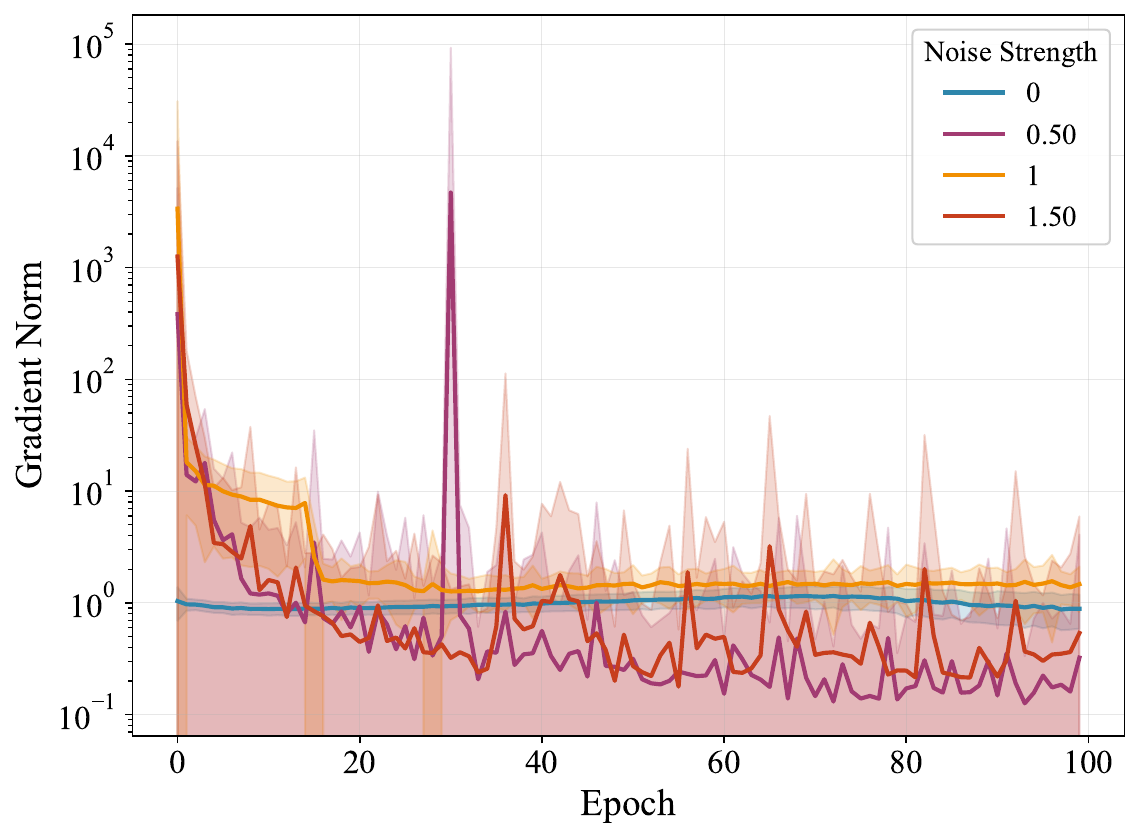}
\caption{Strongly coupled}
\label{fig:grad-strongly-coupled}
\end{subfigure}
\hfill
\begin{subfigure}{0.32\textwidth}
\centering
\includegraphics[width=\linewidth]{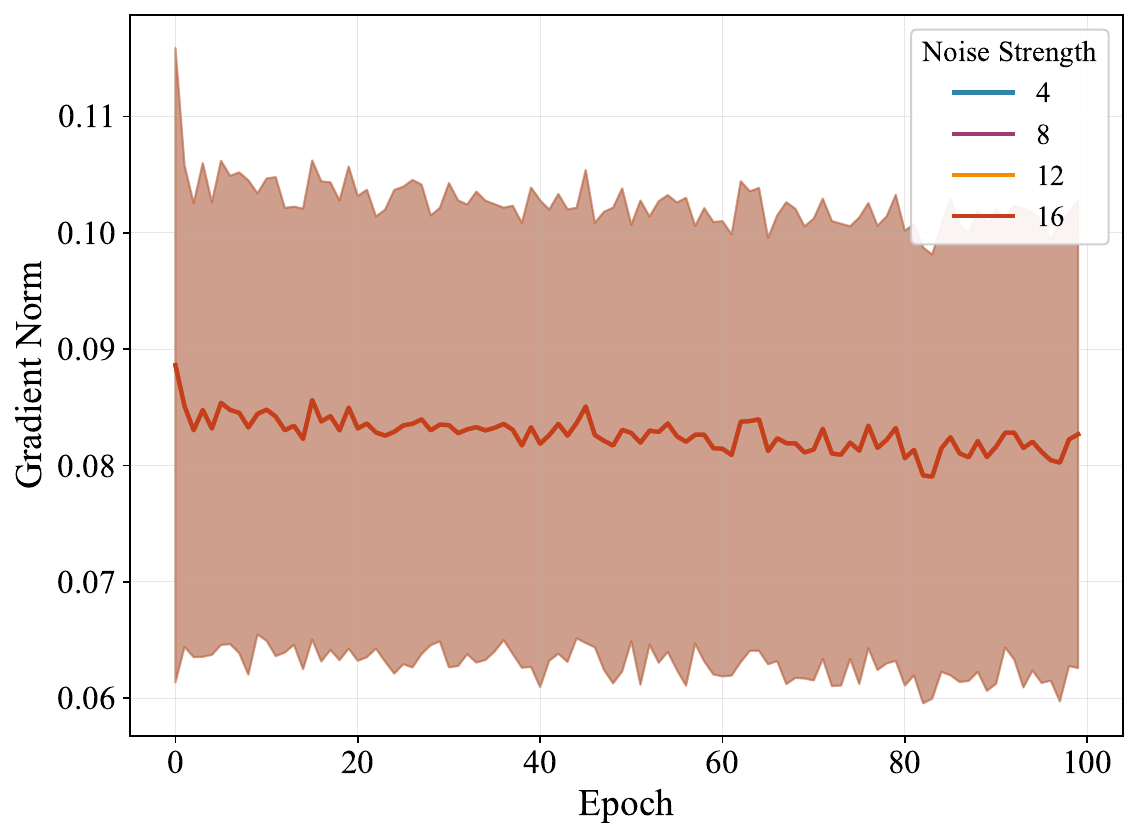}
\caption{Discretization}
\label{fig:grad-discretization}
\end{subfigure}
\caption{\textbf{Gradient norm dynamics under six perturbation classes.} Learnable perturbations maintain stable gradient norms, strongly coupled nonlinear perturbations cause large oscillations, and direct discretization collapses task-relevant gradients. Solid lines show gradient norms and shaded regions denote standard deviation across training iterations.}
\label{fig:gradient-dynamics}
\end{figure*}

Figure~\ref{fig:main_figure} summarizes task-level behavior and Figure~\ref{fig:gradient-dynamics} summarizes optimization-level behavior. Together, they reveal three recurring compensation regimes under HAT.

\textbf{Regime I: Fully compensable perturbations.}
Additive perturbations, multiplicative perturbations, and simplified input-dependent structured scaling remain compensable across the tested regimes. Although these perturbations can strongly distort forward inference, they preserve stable gradient dynamics and maintain all three diagnostic patterns to a sufficient degree. In these cases, HAT successfully adapts the trainable weights and recovers near-ideal performance.

\textbf{Regime II: Conditionally compensable perturbations.}
Projection perturbations occupy an intermediate regime. At moderate fault rates, HAT remains effective and gradient norms stay stable, but compensation depends on whether the remaining trainable subspace retains sufficient redundancy. In this case, expectation consistency and variance remain well behaved within the active subspace, while sensitivity is lost on frozen coordinates. Projection-like faults are therefore not universally benign, but they need not immediately break optimization.

\textbf{Regime III: Non-compensable perturbations under the tested vanilla gradient-based training protocol.}
Strongly coupled nonlinear perturbations and direct discretization consistently fail, but for different reasons. Strong nonlinear coupling leads to unstable and highly variable gradients, breaking optimization through variance explosion and non-stationary updates. Direct discretization, by contrast, collapses task-relevant gradients almost everywhere, violating sensitivity even when the forward perturbation itself is deterministic. These perturbations fall outside the regime where the tested vanilla HAT protocol works reliably, suggesting the need for surrogate gradients, calibration, or hardware-side mitigation depending on the distortion type.

\subsection{Interpretation by Perturbation Class}

\textbf{Additive: read noise / sense amplifier noise.}
Additive perturbations are fully compensable in our experiments. While uncompensated inference degrades sharply with increasing read noise, HAT maintains near-ideal accuracy and stable gradient norms. The additive form preserves gradient expectation consistency, induces bounded gradient variance for finite noise strength, and retains non-degenerate sensitivity. In this regime, the perturbation acts like structured stochastic gradient noise rather than an optimization-breaking distortion.

\textbf{Multiplicative perturbations: programming variability / retention drift.}
Multiplicative perturbations also remain compensable across the tested variability and drift regimes. Their gradient dynamics closely resemble the additive case, showing that large forward distortion alone does not imply optimization failure. Because the perturbation preserves a simple scaling relationship with the weights, expectation consistency and bounded variance remain intact, and sensitivity is preserved as long as the scaling operator is non-zero almost surely.

\textbf{Projection: stuck-at / dead-cell / write-failure induced freezing.}
Projection perturbations are conditionally compensable. At moderate fault rates, HAT remains effective and gradient norms stay stable, but successful compensation depends on whether the surviving trainable subspace retains sufficient redundancy. Expectation consistency and variance remain well behaved within the active subspace, whereas sensitivity is violated on frozen coordinates. This explains why projection perturbations reduce adaptation capacity without necessarily destabilizing optimization.
In this paper, redundancy is treated qualitatively as the availability of remaining degrees of freedom for reconfiguration, rather than as a closed-form threshold derived for all architectures.

\textbf{Input-dependent structured scaling: simplified IR-drop for large crossbar arrays.}
Input-dependent structured scaling remains learnable when the input dependence is mediated through low-order statistics rather than strong sample-specific coupling. Empirically, simplified IR-drop-style attenuation preserves near-ideal compensated accuracy and stable gradients. The perturbation remains approximately linear at the operator level, so expectation consistency is only mildly biased, gradient variance remains controlled, and sensitivity is preserved.

\textbf{Strongly coupled nonlinear: high-fidelity IR-drop / parasitic coupling models.}
Strongly coupled nonlinear perturbations consistently cause optimization collapse. This does not rule out specialized training methods, hardware-in-the-loop adaptation, or calibrated compensation; rather, it indicates that direct forward-perturbation HAT is insufficient in this regime. Both compensated and uncompensated models fail, and gradient norms exhibit large oscillations and extreme variability. This regime most clearly violates the diagnostic framework: higher-order coupling breaks expectation consistency, induces severe gradient instability in practice, and makes usable gradient signals highly erratic across batches.


\textbf{Discretization: ADC / low-precision readout path.}
Direct discretization provides a complementary failure mode. Here optimization does not fail because of exploding variance, but because task-relevant gradients vanish almost everywhere. The resulting gradient collapse violates non-degenerate sensitivity even when the forward perturbation itself is deterministic. The recovery of training under STE therefore acts as a diagnostic intervention in this setup: restoring gradient accessibility can recover optimization, but this should not be interpreted as a complete evaluation of quantization-aware training methods.

Direct discretization is also diagnostically useful because it separates gradient accessibility from practical compensability~\cite{yin2019understanding}. Although surrogate-gradient methods such as STE can restore backward sensitivity, restoring differentiability alone does not necessarily yield stronger compensation unless quantization materially alters the optimization trajectory.

\subsection{Summary}

Table~\ref{tab:learnability-summary} summarizes the diagnostic status of the six perturbation classes. The key pattern is structural rather than severity-based: perturbations that preserve expectation consistency, variance control, and gradient accessibility remain compensable under HAT, whereas strongly coupled or non-smooth perturbations do not. Projection-like faults occupy an intermediate regime, where compensation depends on available redundancy.

\begin{table}[htbp]
\small
\centering
\begin{tabular}{p{3.2cm}p{1.1cm}p{1.1cm}p{1.2cm}}
\hline
\textbf{Perturbation Class} & \textbf{Exp. Consistency} & \textbf{Var. Control} & \textbf{Sensitivity} \\ 
\hline
Additive & \checkmark & \checkmark & \checkmark \\
Multiplicative (Scaling) & \checkmark & \checkmark & \checkmark \\
Projection (Freezing) & \checkmark & \checkmark & \textasteriskcentered \\
Input-Dependent Scaling & \checkmark & \checkmark & \checkmark \\
Strongly Coupled Nonlinear & \textasteriskcentered & $\times$ & $\times$ \\
Discretization (Direct) & \checkmark & \checkmark & $\times$ \\ \hline
\end{tabular}
\caption{\textbf{Diagnostic summary of the six perturbation classes under the three learnability signals.} "\textasteriskcentered" denotes conditionally satisfied, depending on parameter redundancy or coupling strength.}
\label{tab:learnability-summary}
\end{table}

\section{Implications for Hardware-Aware Training and Hardware Design}
\label{sec:implications}

Our results suggest that hardware non-idealities should not be treated as interchangeable noise sources during HAT. Instead, their mitigation strategy should be determined by whether the induced perturbation remains compatible with gradient-based optimization.

\begin{table*}[t]
\centering
\small
\begin{tabular}{p{0.24\linewidth} p{0.24\linewidth} p{0.22\linewidth} p{0.22\linewidth}}
\toprule
\textbf{Perturbation structure} & \textbf{Observed diagnostic symptom} & \textbf{HAT suitability} & \textbf{Suggested mitigation} \\
\midrule
Additive / multiplicative noise & Stable gradient norms and bounded variability & Suitable for vanilla HAT & Training-time injection, noise-aware training \\
Input-dependent structured scaling & Mildly biased but stable gradients & Usually suitable when mediated by low-order statistics & Simplified IR-drop-aware HAT, array-aware modeling \\
Projection / stuck-at faults & Sensitivity lost on frozen coordinates & Conditional on parameter redundancy & Redundancy, pruning-aware training, fault mapping \\
Strongly coupled nonlinear distortion & Large gradient oscillation and variance amplification & Risky for vanilla HAT & Circuit mitigation, array partitioning, calibration, hardware-in-the-loop adaptation \\
Direct discretization & Task-relevant gradient collapse & Unsuitable without surrogate gradients & STE/QAT, calibration, readout precision adjustment \\
\bottomrule
\end{tabular}
\caption{Practical interpretation of the observed diagnostic regimes under vanilla forward-perturbation HAT.}
\label{tab:practical-guidance}
\end{table*}

\subsection{Training-compatible and hardware-limited non-idealities}

Additive perturbations, multiplicative perturbations, and weak input-dependent structured scaling are generally compatible with offline training because they preserve sufficiently stable gradient signals. For these perturbation classes, HAT can serve as an effective software-side mitigation mechanism even when uncompensated inference degrades substantially.

Projection-like faults are only conditionally compensable. They do not necessarily destabilize optimization, but they reduce the effective trainable subspace. Their recoverability therefore depends on available redundancy in the model or hardware mapping, suggesting that some fault classes may be partly delegated to training only when sufficient capacity remains for reconfiguration.

\subsection{When mitigation must shift beyond training}

Strongly coupled nonlinear perturbations and direct discretization lie outside the compensation regime of vanilla HAT. In the former case, higher-order coupling induces unstable and highly variable gradients; in the latter, task-relevant gradients become inaccessible without auxiliary approximations such as STE. These failure modes indicate that certain non-idealities must be addressed through circuit-level suppression, calibration, architectural redundancy, or explicit surrogate-gradient design rather than by training alone.

More broadly, our results highlight that perturbation structure matters more than perturbation magnitude alone. A perturbation may severely distort forward inference yet remain compensable if it preserves stable optimization signals, while a more physically detailed model may become less trainable if it introduces strong nonlinear coupling. This suggests that trainability should be treated as a co-design constraint when modeling and mitigating hardware non-idealities in emerging AI accelerators.
This does not diminish the value of detailed hardware modeling for circuit validation; rather, it highlights that modeling fidelity and training compatibility are distinct design objectives.

A simplified perturbation model may be more useful than a higher-fidelity one if the latter introduces strong coupling that HAT cannot effectively absorb.

\subsection{Design implication}
For hardware non-idealities that preserve stable and accessible gradients, training-time compensation is a reasonable first-line strategy. For perturbations that destroy gradient stability or accessibility, training should not be assumed sufficient, and hardware-side mitigation becomes necessary. This perspective suggests that trainability itself should be treated as a design constraint when selecting hardware models and compensation strategies.

\subsection{Scope and outlook}
Although the present study is controlled and architecture-specific, the diagnostic perspective is useful precisely because it isolates operator-level properties that can recur across different hardware instantiations. In this sense, the framework is intended less as an architecture-specific benchmark and more as a compact way to reason about whether a non-ideality is likely to remain compatible with gradient-based compensation. An important next step is to extend this diagnostic view beyond offline HAT to settings with online calibration, hardware-in-the-loop adaptation, or architecture-aware redundancy allocation, where mitigation can be distributed more dynamically across software and hardware.

\section{Conclusion}

We presented a diagnostic framework for understanding when hardware-aware training can compensate hardware-induced perturbations, and showed that compensability is governed primarily by optimization compatibility rather than perturbation magnitude alone. These results provide a practical co-design perspective: non-idealities that preserve stable and accessible gradients can often be delegated to training, whereas those that destroy gradient stability or accessibility require hardware-side mitigation.
We hope this controlled diagnostic viewpoint provides a useful starting point for deciding when vanilla HAT is a reasonable mitigation strategy and when more specialized training, calibration, or hardware-side intervention should be considered.

Code is available at: \url{https://github.com/MiCrSYZ/HW-aware_training.git}

\bibliographystyle{ACM-Reference-Format}
\bibliography{GLSVLSI/ref}

\newpage
\appendix
\onecolumn
\section{Appendix}

\subsection{Hardware-Aware Training (HAT) Formulation and Algorithm}
\label{app:hat-formulation}

The core idea of HAT is to inject simulated hardware non-idealities into the forward pass during training, enabling the network to learn to compensate for these imperfections and maintain performance under physical hardware constraints.

\subsubsection{Mathematical Formulation}
Given an ideal weight matrix $W\in\mathbb{R}^{m \times n}$, we first clamp it to the programmable range:
\[
W_{\text{clamp}} = \mathrm{clip}(W, W_{\min}, W_{\max}),
\]
then decompose it into positive and negative components:
\[
\begin{cases}
W_p = \frac{\max(W_{\text{clamp}}, 0)}{\max_{i,j} |W_{\text{clamp},ij}|}, \\
W_n = \frac{\max(-W_{\text{clamp}}, 0)}{\max_{i,j} |W_{\text{clamp},ij}|}.
\end{cases}
\]

These are mapped to conductance values $G_p,G_n\in [G_{\min},G_{\max}]^{m\times n}$:
\[
\begin{cases}
G_p = G_{\min} + W_p(G_{\max} - G_{\min}), \\
G_n = G_{\min} + W_n(G_{\max} - G_{\min}).
\end{cases}
\]

Hardware non-idealities are then injected independently into $G_p$ and $G_n$:
\[
\tilde G_p = \mathcal{F}(G_p; t, \xi_p), \quad \tilde G_n = \mathcal{F}(G_n; t, \xi_n),
\]
where $\mathcal{F}$ is the non-ideality operator, $t$ is the time/cycle index, and $\xi_p,\xi_n$ are random variables representing stochastic effects like device variability and read noise.

The effective weight matrix is reconstructed from the perturbed conductances:
\[
W_{\text{eff}} = (\tilde G_p - \tilde G_n) \cdot \frac{\max \lVert W \rVert}{G_{\max}-G_{\min}}.
\]

The forward pass using the perturbed weights is:
\[
y = f(x; W, \xi, t) = W_{\text{eff}} x + b,
\]
and the training objective minimizes the expected loss over both data and hardware perturbations:
\[
\mathcal{L}_t(W) = \mathbb{E}_{(x,y)\sim\mathcal{D}} \mathbb{E}_{\xi\sim\Xi}\Big[ \ell\big(f(x;W,\xi,t),y\big) \Big] + \lambda_{\text{reg}} \cdot \mathcal{L}_{\text{reg}}(W),
\]
where $\mathcal{D}$ is the data distribution, $\Xi$ is the distribution of hardware perturbations, $\ell$ is the task loss (e.g., cross-entropy), and $\mathcal{L}_{\text{reg}}$ is a regularization term that encourages weights to stay away from the clipping boundaries to improve hardware robustness.

\subsubsection{Algorithmic Procedure}
The complete HAT procedure is summarized in Algorithm~\ref{alg:hat}, which integrates the above formulation into an end-to-end training loop.

\begin{algorithm}[t]
\caption{Hardware-Aware Training (HAT)}
\label{alg:hat}
\begin{algorithmic}[1]
\State \textbf{Input:} Training dataset $\mathcal{D}_{\text{train}}$, initial weights $W^{(0)}$, learning rate $\eta$
\State \hspace{1.6em} (Optional) range regularization parameters $\lambda_{\text{reg}}$, $\beta$
\State \textbf{Initialize:} iteration counter $k \leftarrow 0$
\While{not converged}
    \State Sample a mini-batch $(x, y) \sim \mathcal{D}_{\text{train}}$
    \State Sample hardware non-idealities $\xi \sim \Xi$
    \State Construct effective weights:
    \[
        W_{\text{eff}} = \mathcal{R}\big( \Delta(\mathcal{M}(W^{(k)}); \xi, t) \big)
    \]
    \State Forward propagation: $\hat{y} = f(x; W_{\text{eff}})$
    \State Compute task loss: $\ell_{\text{task}} = \ell(\hat{y}, y)$
    \State Initialize total loss: $\mathcal{L} \leftarrow \ell_{\text{task}}$
    \If{range regularization is enabled}
        \State Compute regularization loss:
        \[
            \mathcal{L}_{\text{reg}} =
            \frac{1}{L} \sum_{l=1}^{L} \frac{1}{N_l}
            \sum_{i=1}^{N_l}
            \left[\max\big(|W_i^{(l)}| - \beta W_{\max},\, 0\big)\right]^2
        \]
        \State Update total loss:
        \[
            \mathcal{L} \leftarrow \mathcal{L}
            + \lambda_{\text{reg}} \cdot \mathcal{L}_{\text{reg}}
        \]
    \EndIf
    \State Backpropagation:
    \[
        g = \nabla_{W^{(k)}} \mathcal{L}
    \]
    \State Update weights:
    \[
        W^{(k+1)} = W^{(k)} - \eta \cdot g
    \]
    \State $k \leftarrow k + 1$
\EndWhile
\State \textbf{Output:} trained weights $W^{(k)}$
\end{algorithmic}
\end{algorithm}

\subsection{Gradient Expectation Consistency}
\label{sec:Gradient Expectation Consistency}

In Section~\ref{sec:learnability-boundary}, we employ the approximation
\[
\nabla_W \mathbb{E}_\xi[\ell(f(x; W, \xi))] \approx \mathbb{E}_\xi[\nabla_W \ell(f(x; W, \xi))],
\]
which we refer to as gradient expectation consistency. This approximation is standard in stochastic optimization and hardware-aware training, but we briefly clarify its scope and limitations here.

From a theoretical perspective, exchanging the gradient and expectation operators requires regularity conditions such as local Lipschitz continuity of the loss with respect to (W), finite moments of the stochastic gradients, and independence between the sampled perturbations $\xi$ and the model parameters. These conditions are not guaranteed globally for deep neural networks, especially in the presence of non-smooth activations and batch-dependent operations.

In our setting, the injected hardware non-idealities are sampled independently of the network parameters and are treated as constants during backpropagation. That is, gradients are not propagated through the noise generation process itself. Under this implementation choice, the stochastic gradient computed during training is an unbiased estimator of the gradient of the expected loss, making the above approximation exact for the executed optimization procedure.

We emphasize that this assumption is not intended as a strong theoretical guarantee, but rather as a modeling abstraction that allows us to focus on how different structural properties of hardware non-idealities affect optimization dynamics. Violations of gradient expectation consistency, such as those induced by strongly coupled or non-stationary perturbations, are precisely the regimes where learnability breaks down, as analyzed in the main text.

\subsection{Lipschitz Continuity of Optimal Solutions under Drift}
\label{sec:lipschitz-optimal}

Conductance drift is a systematic time-dependent attenuation of conductance values. We model it as:
\[
G_t = G_0 \cdot \left(1 - \alpha \log\left(1 + \frac{t}{\tau}\right)\right),
\]
where $\alpha$ is the drift coefficient and $\tau$ is a time constant (set to $\tau=1$ in this study).

For a fixed weight $W$, the loss function $\mathcal{L}_t(W)$ is Lipschitz continuous in time $t$:
\[
\exists L_t > 0, \quad \forall t_1, t_2: \quad 
\big| \mathcal{L}_{t_1}(W) - \mathcal{L}_{t_2}(W) \big|
\le L_t |t_1 - t_2|.
\]

Different times $t$ correspond to a family of loss functions $\{\mathcal{L}_t\}_{t\geq0}$ with similar shapes but slowly shifting optimal points:
\[
W_t^* = \arg\min_W \mathcal{L}_t(W).
\]

Under assumptions of strong convexity and Lipschitz continuity of $\nabla_W \mathcal{L}_t(W)$ with respect to both $W$ and $t$, the trajectory of optimal solutions satisfies:
\[
\| W_{t_1}^* - W_{t_2}^* \| \le C |t_1 - t_2|,
\]
where $C$ is a constant depending on the convexity and smoothness parameters. This Lipschitz property ensures that the optimal weight configuration evolves slowly over time, making it trackable through expectation-based training over a distribution of time steps.

We note that the following analysis assumes local strong convexity, which does not strictly hold for deep neural networks. This derivation is intended as an idealized analysis to provide intuition on the continuity of optimal solutions under slow hardware drift, rather than as a formal guarantee.
This intuition aligns with empirical observations from online and time-varying optimization, where slow parameter drift often leads to trackable solution paths~\cite{bonnans2013perturbation, simonetto2020time}. Our HAT formulation, which samples $t$ uniformly during training, effectively learns an expectation over this slowly evolving family of objectives, thereby compensating for the drift.

\subsection{Quantitative Relation between Fault Rate and Parameter Redundancy}
\label{sec:fault-redundancy-relation}

Stuck-at faults permanently freeze a subset of parameters, effectively reducing the trainable parameter space. Let $\rho$ be the fault rate, and let the original parameter space be $\mathbb{R}^D$. After faults, the trainable subspace has dimension $\|\mathbf{S}\|_0 = (1-\rho)D$.

Assuming the loss function $\mathcal{L}(W)$ is Lipschitz continuous with constant $L_W$:
\[
|\mathcal{L}(W + \Delta W) - \mathcal{L}(W)| \le L_W \lVert \Delta W \rVert,
\]
we can bound the accuracy degradation $\Delta \text{Acc}$ due to stuck-at faults by controlling the perturbation norm $\lVert \Delta W_{\text{stuck}} \rVert$.

To analyze the effect of parameter redundancy, consider representing a single ideal weight $w$ by $r$ redundant weights $w^{(1)}, \dots, w^{(r)}$, with the effective weight given by their average:
\[
\bar w = \frac{1}{r} \sum_{j=1}^r w^{(j)}.
\]

Suppose each redundant weight $w^{(j)}$ is subject to a stuck-at perturbation $e_j$, which equals $\alpha w$ with probability $p$ (fault) and $0$ otherwise. The mean perturbation is $\mathbb{E}[e_j] = p\alpha w$, and the variance of the average perturbation is:
\[
\mathrm{Var}(\Delta \bar w) = \frac{p(1-p)(\alpha w)^2}{r}.
\]

Extending to the entire weight matrix $W$, the perturbation norm scales as:
\[
\lVert \Delta W_{\text{stuck}} \rVert \sim \sqrt{\frac{p(1-p)}{r}} \lVert W \rVert.
\]

To keep the accuracy drop within a threshold $\varepsilon$, we require:
\[
L_W \cdot C \sqrt{\frac{p(1-p)}{r}} \lVert W \rVert \le \varepsilon,
\]
which yields the necessary redundancy factor:
\[
r \ge p(1-p) \left( \frac{L_W C \lVert W \rVert}{\varepsilon} \right)^2,
\]
where constant $C$ characterizes the worst-case amplification of weight distortion induced by stuck-at faults under a given conductance-to-weight mapping. It explicitly couples device-level non-idealities with algorithmic weight constraints.

This relation quantifies how parameter redundancy can absorb stuck-at faults while maintaining performance, providing a guideline for designing fault-tolerant architectures.

\subsection{Gradient Analysis for Discretization Without STE}
\label{sec:app-boundary-loss}

In experiments with non-differentiable discretization operators (e.g., uniform quantization without STE), the forward pass is $z = Q(Wx)$ where $\partial Q/\partial z = 0$ almost everywhere. 
Consequently, the gradient of the main task loss vanishes:
\[
\nabla_W \mathcal{L}_{\text{task}} = \frac{\partial \ell}{\partial Q} \cdot \frac{\partial Q}{\partial z} \cdot \frac{\partial z}{\partial W} \equiv 0.
\]

To prevent numerical instability and to examine gradient behavior, we employ an auxiliary regularization loss:
\[
\mathcal{L}_{\text{reg}}(W) = \frac{1}{N} \sum_{i=1}^{N} \left(\max(|W_i| - \beta \cdot W_{\max}, 0)\right)^2,
\]
where $N$ is the number of weight elements, $\beta \in (0,1)$ is a threshold ratio, and $W_{\max}$ is the weight-clipping upper bound. The total loss is:
\[
\mathcal{L}(W) = \mathcal{L}_{\text{task}}(Q(Wx), y) + \lambda \cdot \mathcal{L}_{\text{reg}}(W).
\]

Since $\nabla_W \mathcal{L}_{\text{task}} \equiv 0$, the effective gradient reduces to:
\[
\nabla_W \mathcal{L} \approx \lambda \cdot \nabla_W \mathcal{L}_{\text{reg}},
\]
explaining the non-zero but task-irrelevant gradient norms observed in Figure~\ref{fig:grad-discretization}. 
This confirms that without gradient approximation, discretization operators provide no learnable signal for the primary task, rendering them non-learnable under pure gradient-based optimization.

\subsection{Detailed Distortion Calibration Procedure}
\label{sec:calibration-details}

This section provides a complete description of the distortion calibration protocol.

\subsubsection{Distortion Metric Definitions}

We quantify the strength of an injected perturbation using the relative output distortion:
\begin{equation}
\delta
\triangleq
\mathbb{E}\left[
\frac{\|\tilde{\mathbf{y}} - \mathbf{y}\|_2}{\|\mathbf{y}\|_2 + \epsilon}
\right],
\label{eq:delta_def}
\end{equation}
where $\mathbf{y}$ and $\tilde{\mathbf{y}}$ denote the layer outputs before and after IR-drop injection, respectively, and $\epsilon$ is a small constant for numerical stability.

For networks with multiple memristor-mapped layers, we further define a global distortion metric by aggregating over all such layers:
\begin{equation}
\delta_{\text{global}} =
\frac{\sum_{l} \|\tilde{\mathbf{y}}^{(l)} - \mathbf{y}^{(l)}\|_2}
{\sum_{l} \|\mathbf{y}^{(l)}\|_2 + \epsilon}.
\label{eq:delta_global}
\end{equation}
This metric captures the overall relative deviation of the forward operator induced by hardware non-idealities and serves as a natural scalar proxy for perturbation magnitude.

\subsubsection{Calibration Protocol Details}

Given a pre-trained model checkpoint, we compute the global distortion metric $\delta_{\text{global}}$ (Eq.~\ref{eq:delta_global}) over a fixed subset of $N=512$ training samples. For a given IR-drop model $\mathcal{M}$ with a scalar strength parameter $s$, we aim to find $s^*$ such that:

\[
\delta_{\text{global}}(s^*) \approx \delta_{\text{target}},
\]

where $\delta_{\text{target}} \in \{0.05, 0.01\}$ in our experiments.

We employ a simple bisection-inspired search with $20$ trials per target. Starting from an initial guess $s_0 = 1.0$, each trial $t$ evaluates $\delta_{\text{global}}(s_t)$ and adjusts $s_{t+1}$ as:

\[
s_{t+1} = 
\begin{cases}
s_t / 2 & \text{if } \delta_{\text{global}}(s_t) > \delta_{\text{target}} \\
s_t \times 1.5 & \text{otherwise}.
\end{cases}
\]

To avoid infinite loops, we clip $s_t$ to the range $[10^{-8}, 1.0]$. The search terminates early if $\delta_{\text{global}}(s_t)$ stabilizes within $10\%$ of the target.

\subsection{Gradient Approximation for Non-Differentiable Operators}
\label{sec:gradient-approximation}

This section provides supplementary analysis of gradient approximation techniques for non-differentiable operators, which are referenced as diagnostic cases in the main text.

\subsubsection{Gradient Approximation Methods}
For non-differentiable operators, gradient approximation methods can partially restore effective gradient flow within the HAT framework. The core idea is to construct a differentiable proxy gradient during backpropagation that bypasses the non-differentiable points of the original operator.

\begin{enumerate}
    \item \textbf{Straight-Through Estimator (STE)}: The most common gradient approximation method can be formalized as:
    \[
    \text{Forward:} \quad y_{\text{quant}} = Q(y), \qquad
    \text{Backward:} \quad \frac{\partial y_{\text{quant}}}{\partial y} \triangleq 1,
    \]
    where $Q$ is the quantization function. STE essentially assumes the quantization operation has unity gradient, thereby "passing through" gradients to preceding layers.
    
    \item \textbf{Probabilistic Quantization \& Noise Injection}: Deterministic quantization can be modeled as a stochastic process:
    \[
    y_{\text{quant}} = y + \epsilon, \quad \epsilon \sim \mathcal{U}(-\frac{\Delta}{2}, \frac{\Delta}{2}),
    \]
    where backpropagation proceeds directly with unbiased gradients. This approach often exhibits better stability in low-bit quantization scenarios.
    
    \item \textbf{Smooth Approximation Functions}: Differentiable functions can approximate non-differentiable operators, such as using a sigmoid to approximate a step function:
    \[
    \text{Step}(x) \approx \sigma(\alpha x) = \frac{1}{1+e^{-\alpha x}},
    \]
    where $\alpha$ controls approximation accuracy.
\end{enumerate}

While gradient approximation methods can expand the range of perturbations that are trainable in practice, they often introduce biased or high-variance gradients. As a result, trade-offs between learnability, stability, and final accuracy must be considered in conjunction with the specific task and perturbation regime.

\subsubsection{Detailed Analysis of STE's Limitations}
\label{sec:ste-limitations}

STE successfully makes quantization "learnable" by restoring gradient flow, but its practical utility is bounded by two key factors:

\begin{enumerate}
    \item \textbf{Gradient Bias}: The approximation $\frac{\partial Q}{\partial y} \triangleq 1$ introduces systematic bias:
    \[
    \mathbb{E}[\nabla_{\text{STE}}] \neq \nabla_{\text{real}}, \quad \text{Var}(\nabla_{\text{STE}}) \geq \text{Var}(\nabla_{\text{ideal}}),
    \]
    potentially leading to suboptimal convergence trajectories.
    
    \item \textbf{Noise Intensity Threshold}: At higher bit-widths (e.g., 8-bit), the quantization noise variance $\sigma_q^2$ may fall below a regime where quantization noise meaningfully affects optimization dynamics. In such regimes, models become insensitive to quantization noise, making explicit training-time injection unnecessary.
\end{enumerate}

Experimental comparison of two ADC handling strategies:
\begin{enumerate}
    \item No injection during training, inference-time injection: 89.36\% test accuracy
    \item STE-injected quantization during training, inference-time injection: 89.28\% test accuracy
\end{enumerate}

The negligible performance difference observed at moderate bit-widths indicates that the benefit of restoring gradient flow via STE can be offset by gradient bias. This detailed analysis supports the interpretation in the main text that STE primarily serves as a mechanism for preventing gradient collapse, rather than a guarantee of improved optimization outcomes.

\paragraph{Extreme Low-Bit Regime.}
We further examine the behavior of STE under extreme discretization on CIFAR-100 by reducing the ADC precision to 2 bits. While 8-bit and 4-bit quantization retain stable validation and test accuracy, a sharp performance collapse is observed at 2-bit precision, accompanied by a significant increase in final training loss. 
This behavior is consistent with a breakdown of gradient approximation under severe discretization, where the mismatch between the surrogate STE gradient and the true gradient of the quantized operator becomes dominant. In this regime, restoring gradient flow alone is insufficient to ensure stable optimization, illustrating a concrete failure mode of gradient-based compensation for non-differentiable operators.

\begin{figure*}[h]
\centering
\begin{subfigure}[b]{0.32\textwidth}
\centering
\includegraphics[width=\linewidth]{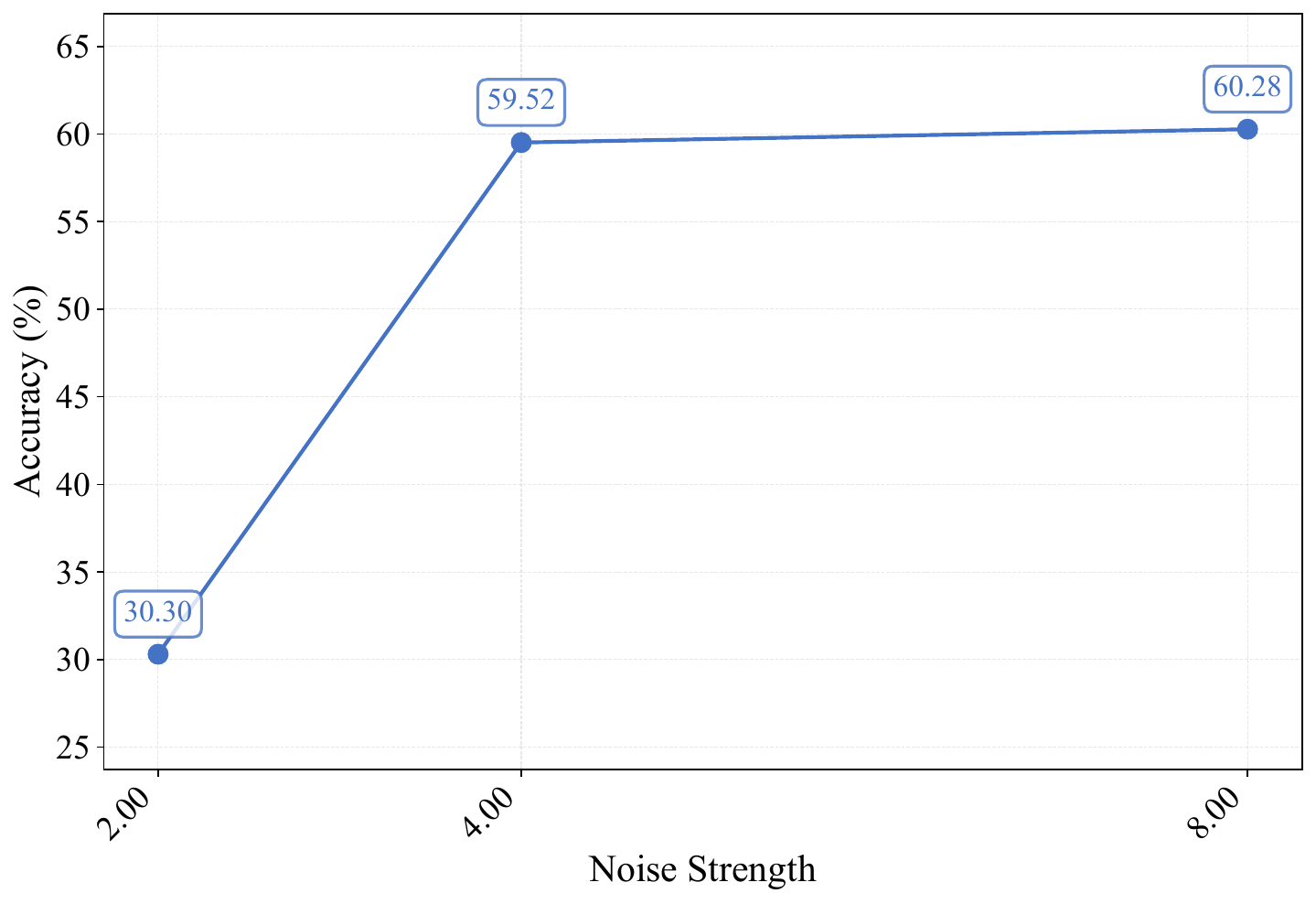}
\caption{Accuracy}
\label{fig:adc_ste_accuracy_cifar100}
\end{subfigure}
\hfill
\begin{subfigure}[b]{0.32\textwidth}
\centering
\includegraphics[width=\linewidth]{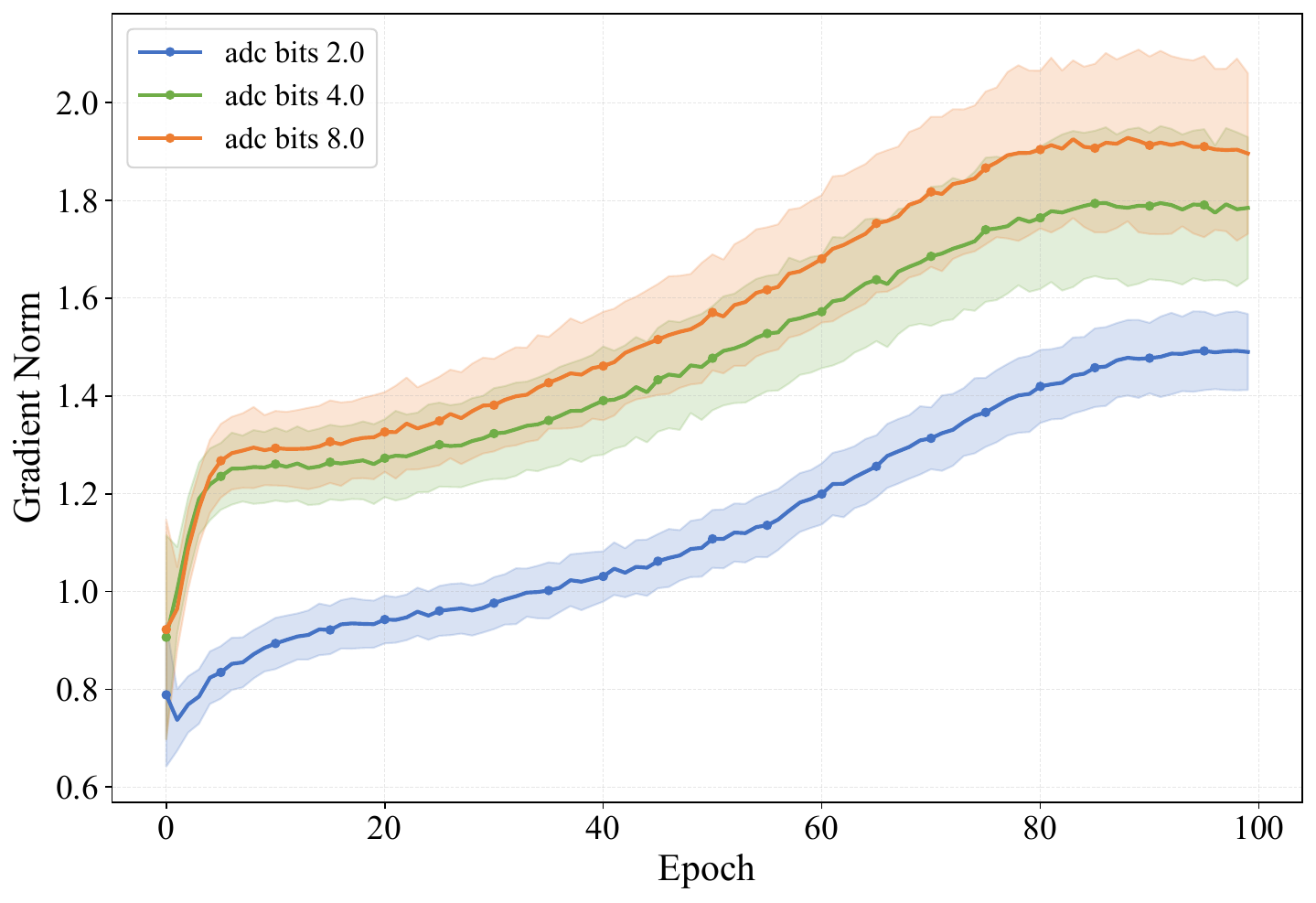}
\caption{Gradient Norm}
\label{fig:adc_ste_grad_cifar100}
\end{subfigure}
\hfill
\begin{subfigure}[b]{0.32\textwidth}
\centering
\includegraphics[width=\linewidth]{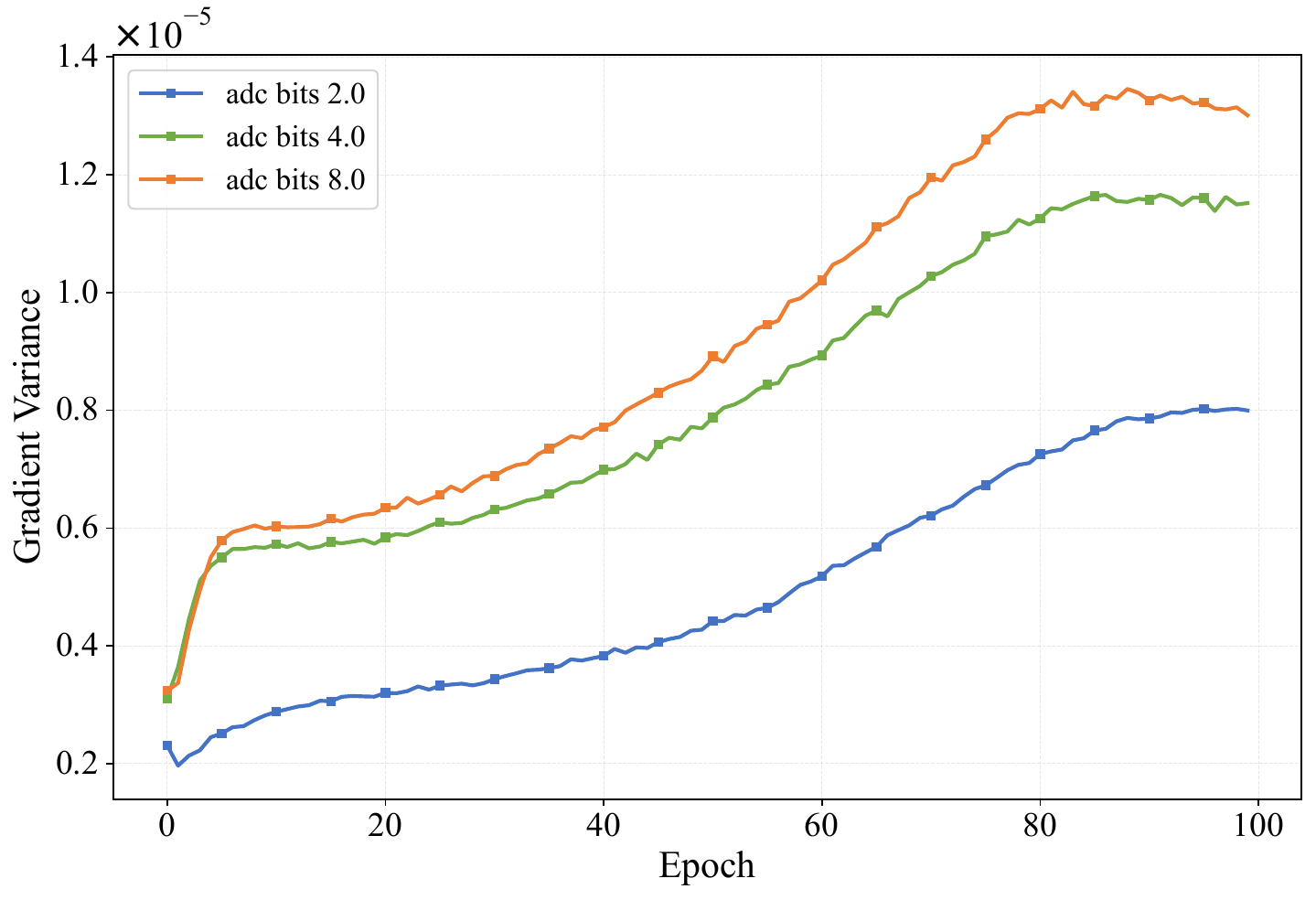}
\caption{Gradient Variance}
\label{fig:adc_ste_grad_var_cifar100}
\end{subfigure}
\caption{Accuracy, gradient norm and gradient variance of STE-based ADC quantization on CIFAR-100 under different bit-widths.}
\label{fig:adc_ste_cifar100}
\end{figure*}

\subsubsection{Higher-Order Effects of Quantization Noise}
This subsection further examines higher-order effects of quantization noise to explain why explicit training-time injection may offer limited benefits at moderate bit-widths.

When quantization bit-width is sufficiently high, the nonlinear component $\eta_{\text{nonlinear}}$ becomes negligible, and:
\begin{itemize}
    \item The ideally trained model already possesses implicit robustness to linear noise through standard training.
    \item The STE-trained model learns explicitly but suffers from gradient bias.
    \item Their performance converges, consistent with the theoretical prediction that linear noise can be adapted to implicitly.
\end{itemize}

\subsection{Accuracy Statistics During Training}
\label{app:acc_stats}

For perturbations exhibiting stable optimization dynamics, the gap between validation and test accuracy is reduced and performance variance across seeds is lower. 
These trends are consistent with the gradient-level diagnostics discussed in the main text.

\begin{figure*}[h]
\centering
\begin{subfigure}[b]{0.42\textwidth}
\centering
\includegraphics[width=\linewidth]{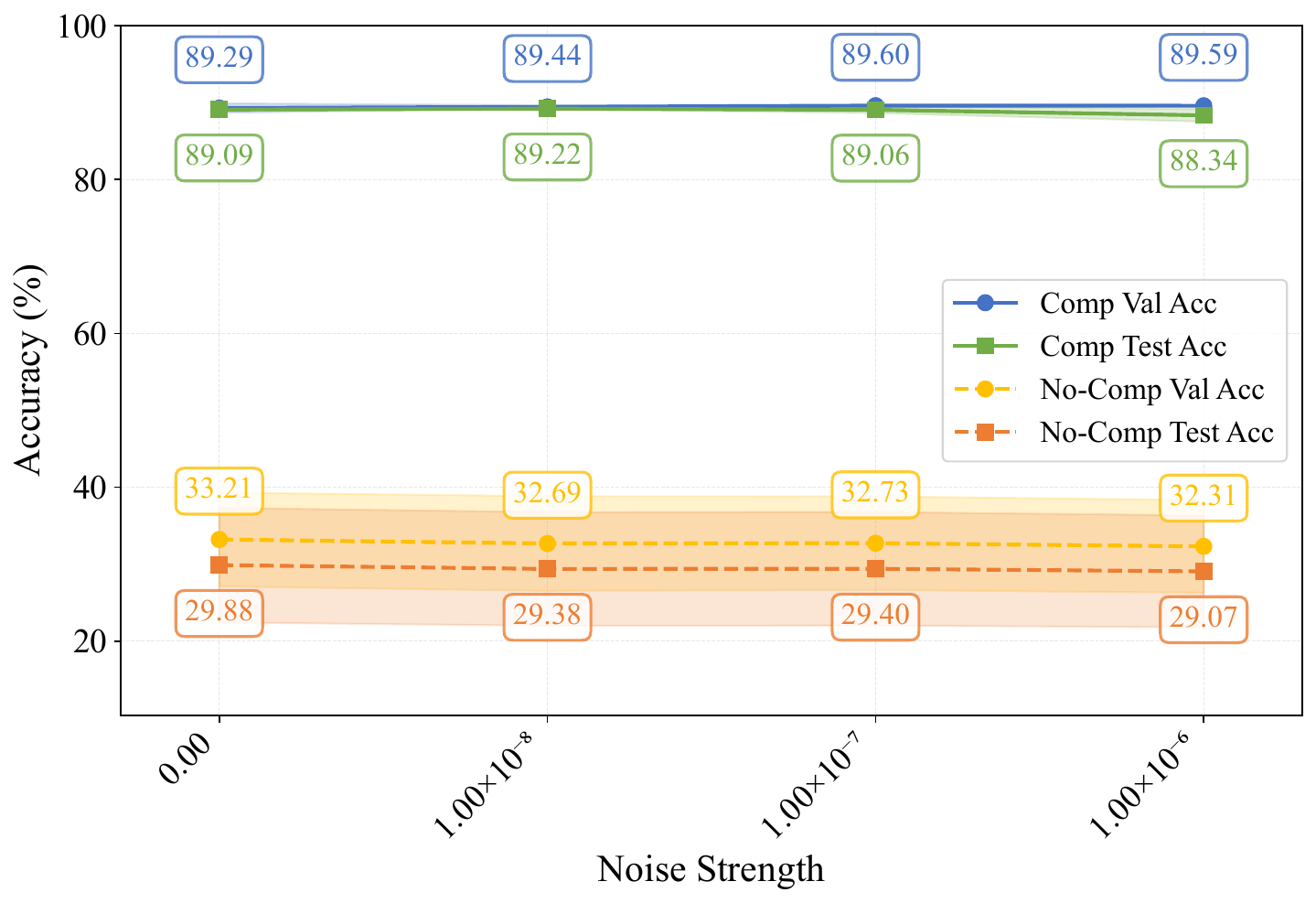}
\caption{Additive: $\sigma_r$}
\label{fig:acc-additive}
\end{subfigure}
\hfill
\begin{subfigure}[b]{0.42\textwidth}
\centering
\includegraphics[width=\linewidth]{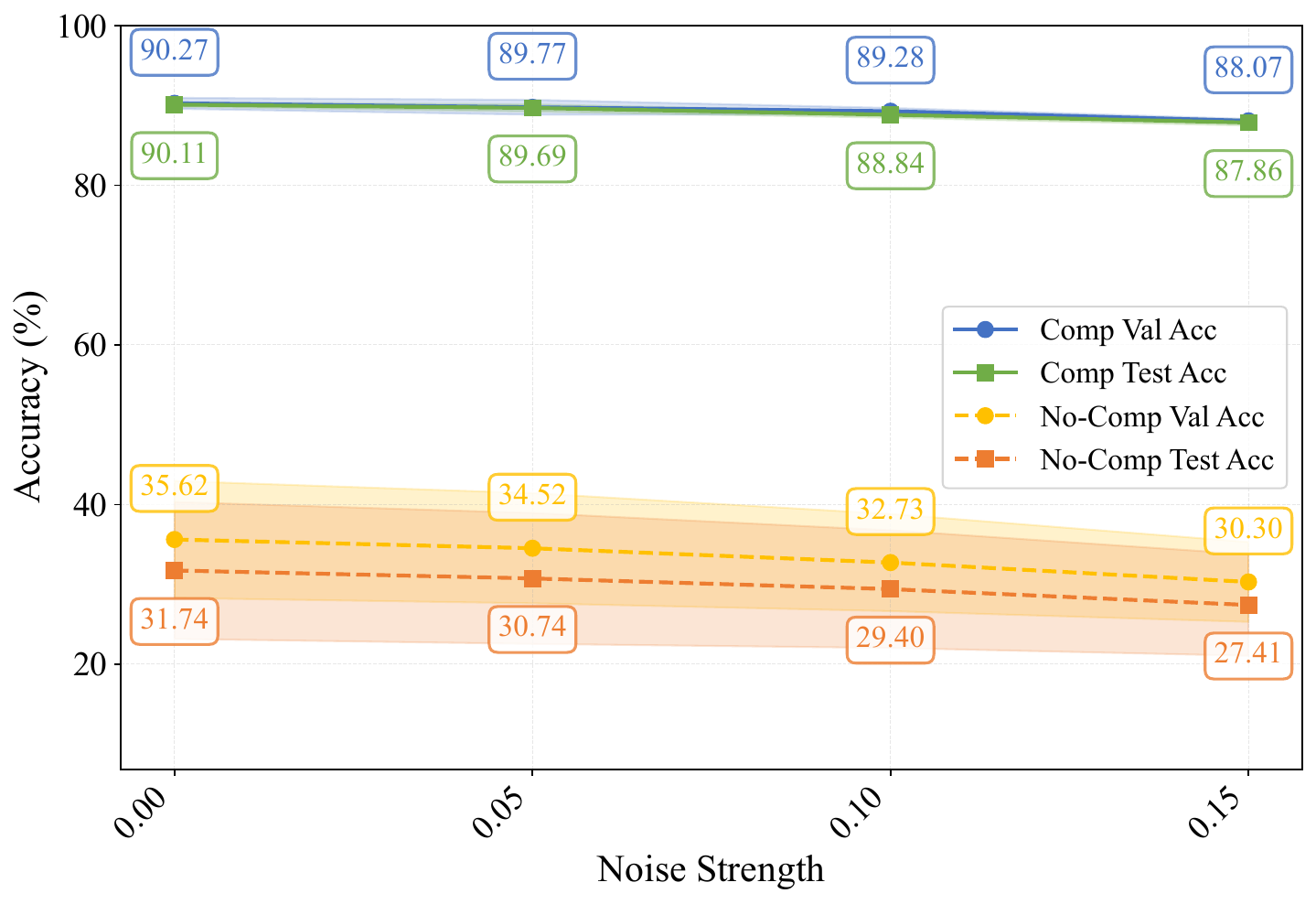}
\caption{Multiplicative: $\sigma_v$}
\label{fig:acc-multiplicative}
\end{subfigure}
\hfill
\begin{subfigure}[b]{0.42\textwidth}
\centering
\includegraphics[width=\linewidth]{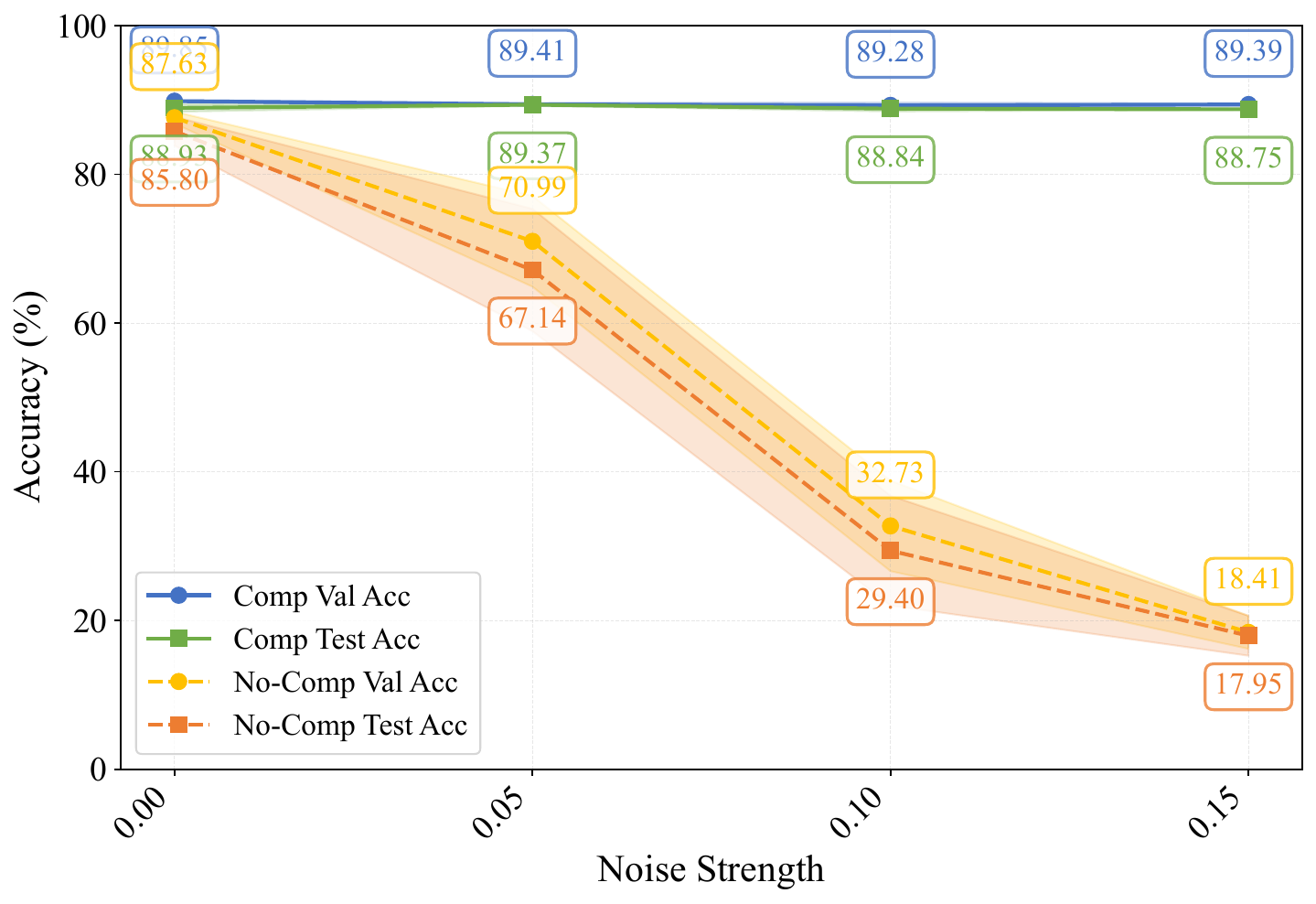}
\caption{Projection: $\rho$}
\label{fig:acc-projection}
\end{subfigure}
\hfill
\begin{subfigure}[b]{0.42\textwidth}
\centering
\includegraphics[width=\linewidth]{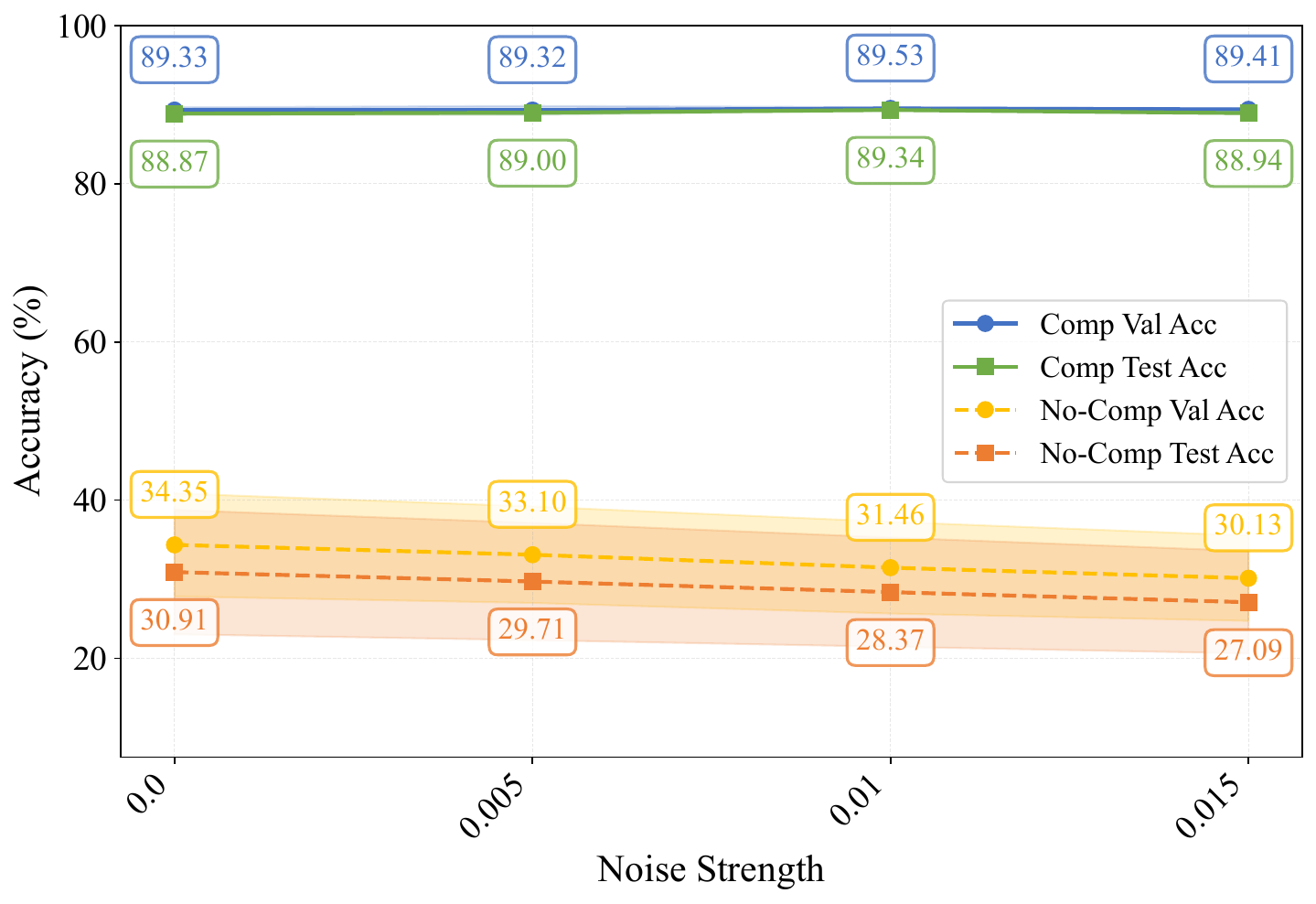}
\caption{Input scaling: $\beta$}
\label{fig:acc-input-scaling}
\end{subfigure}
\caption{
\textbf{Validation and test accuracy statistics under hardware-aware training across learnable perturbation classes.}
(a) Additive perturbations (read noise $\sigma_r$). (b) Multiplicative perturbations (variability $\sigma_v$). (c) Projection perturbations (stuck-at ratio $\rho$). (d) Input-dependent scaling (IR-drop $\beta$).}
\label{fig:acc-statistics}
\end{figure*}

\subsection{Gradient Statistics During Training}
\label{sec:grad_stats}

We report the variance of gradient statistics across training iterations, computed over all model parameters. 
Consistent with the proposed diagnostic framework, perturbations leading to optimization failure are associated with significantly higher variability and instability in gradient behavior.

\begin{figure*}[h]
\centering
\begin{subfigure}[b]{0.32\textwidth}
\centering
\includegraphics[width=\linewidth]{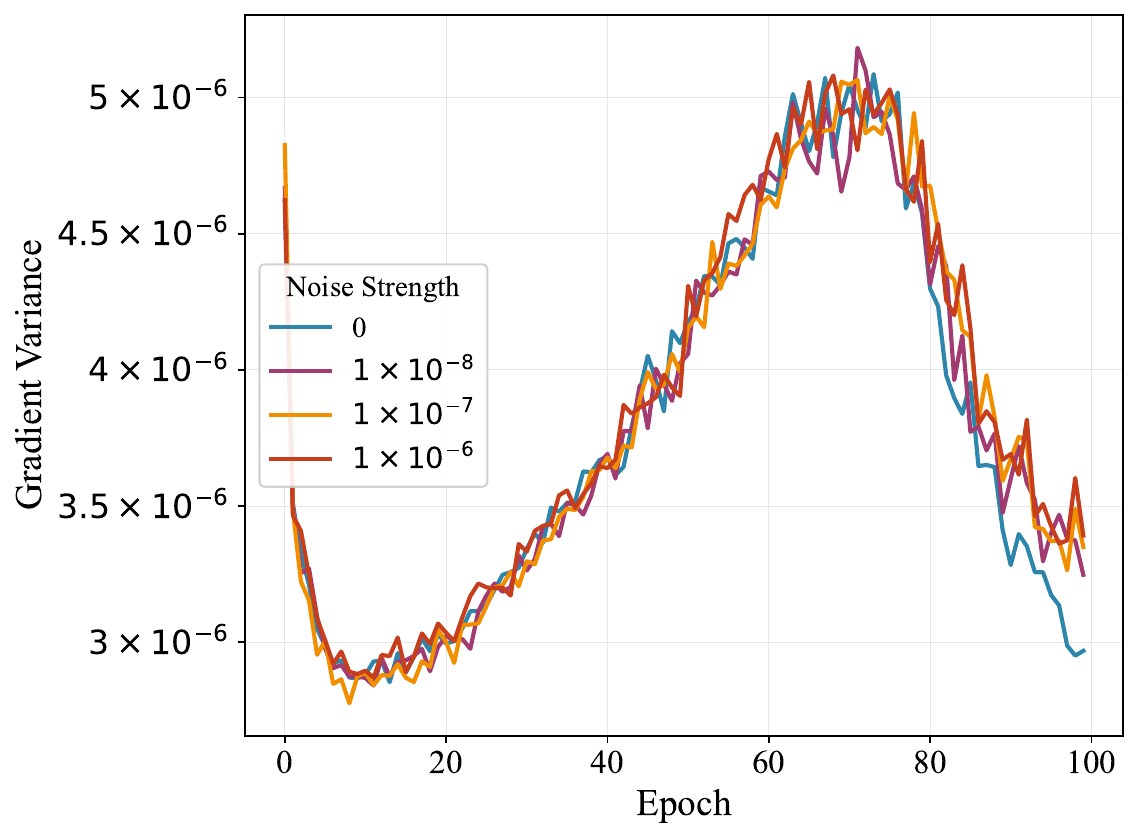}
\caption{Additive: $\sigma_r$}
\label{fig:grad-var-additive}
\end{subfigure}
\hfill
\begin{subfigure}[b]{0.32\textwidth}
\centering
\includegraphics[width=\linewidth]{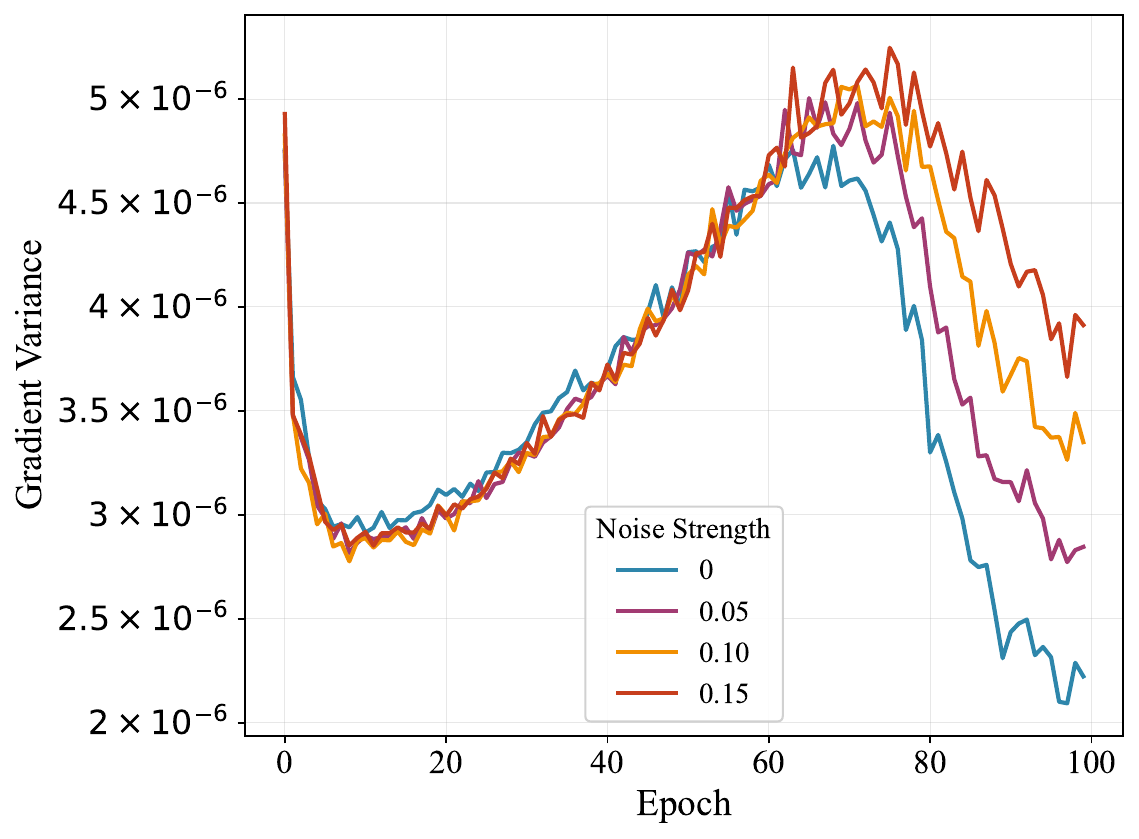}
\caption{Multiplicative: $\sigma_v$}
\label{fig:grad-var-multiplicative}
\end{subfigure}
\hfill
\begin{subfigure}[b]{0.32\textwidth}
\centering
\includegraphics[width=\linewidth]{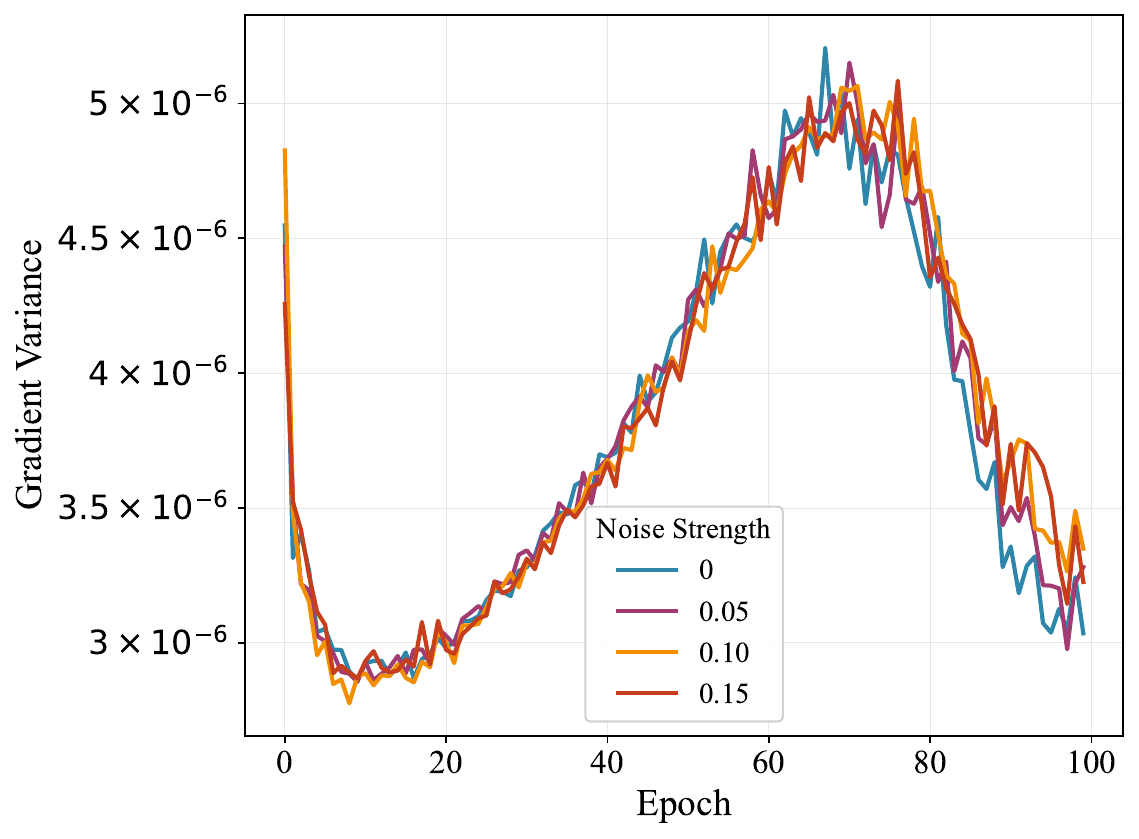}
\caption{Projection: $\rho$}
\label{fig:grad-var-projection}
\end{subfigure}
\hfill
\begin{subfigure}[b]{0.32\textwidth}
\centering
\includegraphics[width=\linewidth]{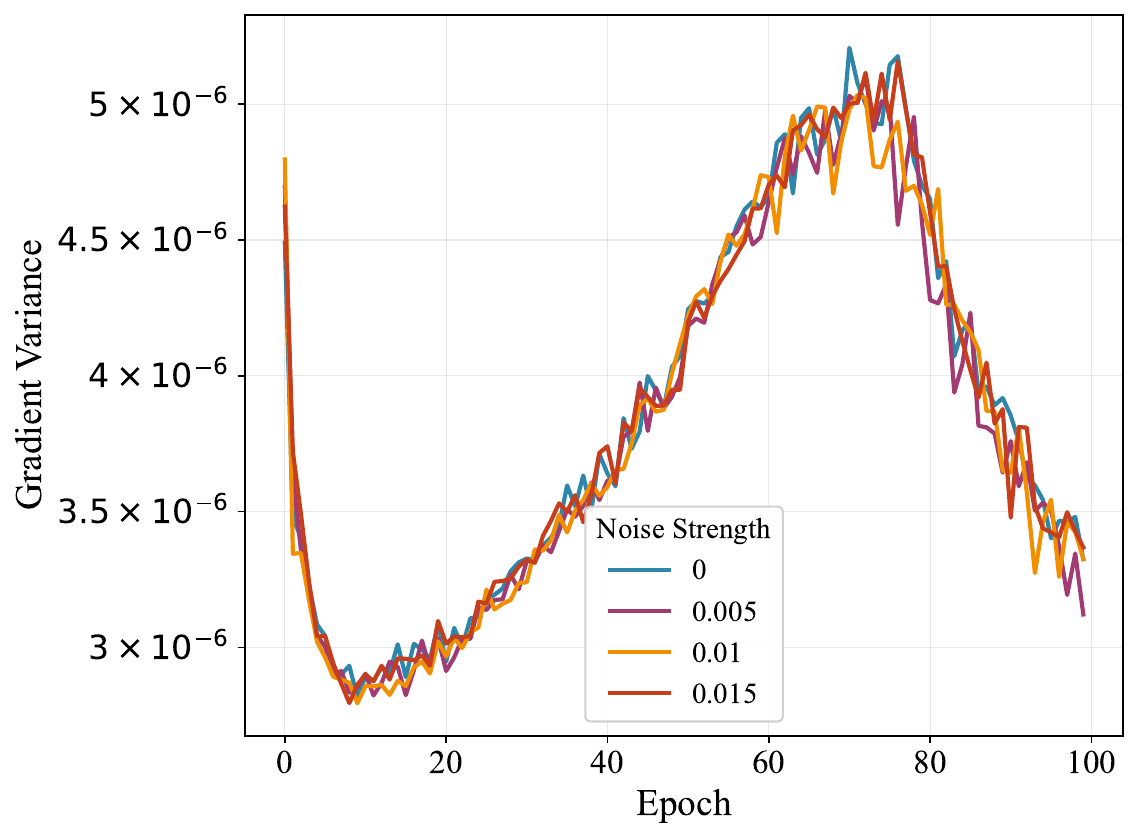}
\caption{Input scaling: $\beta$}
\label{fig:grad-var-input-scaling}
\end{subfigure}
\hfill
\begin{subfigure}{0.32\textwidth}
\centering
\includegraphics[width=\linewidth]{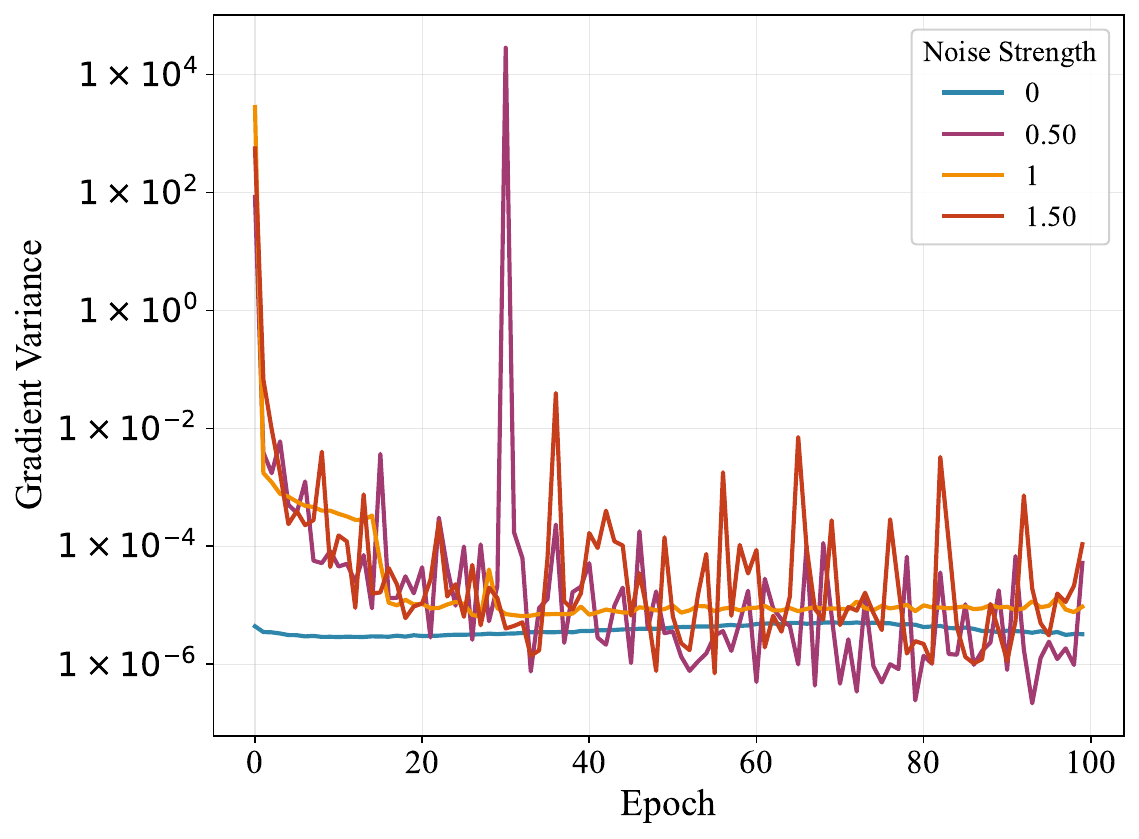}
\caption{Strongly coupled}
\label{fig:grad-var-strongly-coupled}
\end{subfigure}
\hfill
\begin{subfigure}{0.32\textwidth}
\centering
\includegraphics[width=\linewidth]{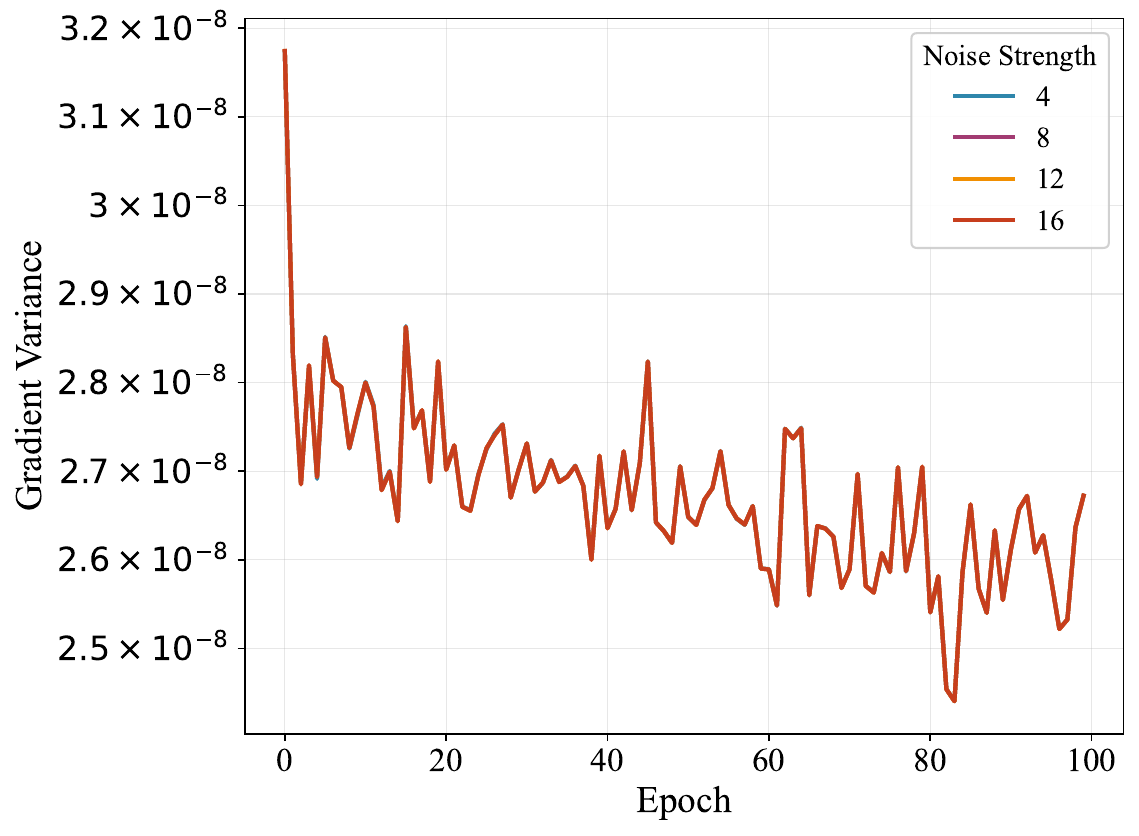}
\caption{Discretization}
\label{fig:grad-var-discretization}
\end{subfigure}
\caption{
\textbf{Gradient variance across perturbation classes with varying strengths.}
(a) Additive perturbations (read noise $\sigma_r$). (b) Multiplicative perturbations (variability $\sigma_v$). (c) Projection perturbations (stuck-at ratio $\rho$). (d) Input-dependent scaling (IR-drop $\beta$). (e) Strongly coupled nonlinear IR-drop. (f) Discretization via ADC quantization.}
\label{fig:gradient-variance}
\end{figure*}

\subsection{Additional Non-Idealities Not Considered in Training}
\label{app:write-variation}

For completeness, we briefly describe additional non-idealities modeled in our simulation framework but not injected during training in this work.

\subsubsection{Cycle-to-Cycle Write Variation}
\label{app:write-variation-detailed}

Cycle-to-cycle write variation (or "write/update model") models stochastic programming errors that occur during repeated write operations in resistive memory arrays. It primarily affects the accuracy of weight programming rather than inference-time computation. Since our analysis focuses on perturbations that affect the forward operator during inference and training, cycle-to-cycle write variation is not included in the main learnability study. Nevertheless, the model is included in our simulator to support future investigations of programming-aware training.

The write process is modeled iteratively until the programmed conductance converges to a target value within a tolerance threshold or a maximum number of iterations is reached:

\[
\begin{aligned}
&\text{for } i = 0, 2, \ldots, m,\ m \le M_{\max}: \\
&\quad \Delta G^{(i)} = 
\begin{cases} 
A_+ \cdot \left(1 - \dfrac{G^{(i-1)}}{G_{\max}}\right)^{p_+} \cdot \left(1-e^{-\gamma V_i t_i}\right) + \xi^{(i)}, & \text{direction}_i = +1, \\
-A_- \cdot \left(\dfrac{G^{(i-1)}}{G_{\max}}\right)^{p_-} \cdot \left(1-e^{-\gamma V_i t_i}\right) + \xi^{(i)}, & \text{direction}_i = -1,
\end{cases} \\
&\quad G^{(i)} = \operatorname{clip}\left(G^{(i-1)} + \Delta G^{(i)},\ G_{\min},\ G_{\max}\right), \\
&\text{until } \lVert G^{(m)} - G_\text{target} \rVert_\infty \le \delta_\text{tolerance}(G_\text{max} - G_\text{min}), 
\end{aligned}
\]

where $M_{\max}$ is the maximum iteration count, $A_+$ and $A_-$ model asymmetric device behavior, and write noise $\xi^{(i)} \sim \mathcal{N}(0,(\sigma_w|\Delta G|)^2)$ captures stochastic programming errors.

\subsection{Experimental Parameters and Baseline Hardware Configuration}
\label{app:experiment-params}

Unless otherwise stated, all experiments in this paper use the default configuration listed in Tables~\ref{tab:training-config} and~\ref{tab:device-config}.

\begin{table}[h]
\centering
\begin{tabular}{l l}
\hline
\textbf{Parameter} & \textbf{Value} \\
\hline
Model architecture & ResNet-20 \\
Dataset & CIFAR-10 / CIFAR-100 \\
Optimizer & SGD \\
Initial learning rate & 0.1 \\
Momentum & 0.9 \\
Weight decay & $1\times10^{-4}$ \\
Batch size & 128 \\
Training epochs & 100 \\
Learning rate schedule & Cosine annealing \\
Loss function & Cross-entropy \\
\hline
\end{tabular}
\caption{Network and training hyperparameters used in all experiments unless otherwise specified.}
\label{tab:training-config}
\end{table}

\begin{table}[h]
\centering
\begin{tabular}{l l}
\hline
\textbf{Parameter} & \textbf{Value} \\
\hline
Conductance range $(G_{\min}, G_{\max})$ & $(1\times10^{-6},\,1\times10^{-4})$ \\
Weight clipping range & $[-1,\,1]$ \\
Mapping function & Linear \\
Array size & 128 \\
ADC resolution & 8-bit \\
ADC enabled during training & False (unless specified) \\
ADC training mode & STE (when enabled) \\
Variability standard deviation $\sigma_v$ & 0.1 \\
Read noise standard deviation $\sigma_r$ & $1\times10^{-7}$ \\
Drift coefficient $\alpha$ & $1\times10^{-4}$ \\
Drift time mode & Accumulated \\
Stuck-at fault ratio & 0.1 \\
IR-drop mode & Crossbar \\
IR-drop strength parameter $\beta$ & 0.01 \\
\hline
\end{tabular}
\caption{Default device-level and perturbation parameters used in hardware-aware training experiments.}
\label{tab:device-config}
\end{table}

\begin{table}[h]
\centering
\begin{tabular}{l l}
\hline
\textbf{Parameter} & \textbf{Value} \\
\hline
$A_{+}$ & $4\times10^{-5}$ \\
$A_{-}$ & $3\times10^{-5}$ \\
$p_{+}$ & 1.0 \\
$p_{-}$ & 1.0 \\
$\gamma$ & 1.0 \\
Write noise ratio $\sigma_w$ & 0.05 \\
Pulse width $t_{\min}$ & $5\times10^{-9}$ \\
Pulse width scale $t_{\mathrm{scale}}$ & $1\times10^{-6}$ \\
Write voltage $V_{\mathrm{write}}$ & 1.2 \\
Write interval & $\texttt{epochs}$ (default: write once after training) \\
Max pulses $M_{\max}$ & 200 \\
Tolerance $\delta_{\mathrm{tolerance}}$ & 0.02 (fraction of $G_{\max}-G_{\min}$) \\
\hline
\end{tabular}
\caption{Cycle-to-cycle write variation model (write/update model) parameters used during inference. This module is not the focus of our analysis, but it is included in the deployed (inference-time) simulation pipeline for completeness (when mentioned).}
\label{tab:write-config}
\end{table}

When the write/update model (detailed in Appendix~\ref{app:write-variation}) is disabled, the parameters in Table~\ref{tab:write-config} are inactive and do not affect training or inference.

\subsection{Supplemental Results}

\subsubsection{Deterministic Drift: Multiplicative (Scaling) Perturbations}
\label{sec:drift}

\begin{figure*}[h]
\centering
\begin{subfigure}[b]{0.32\textwidth}
\centering
\includegraphics[width=\linewidth]{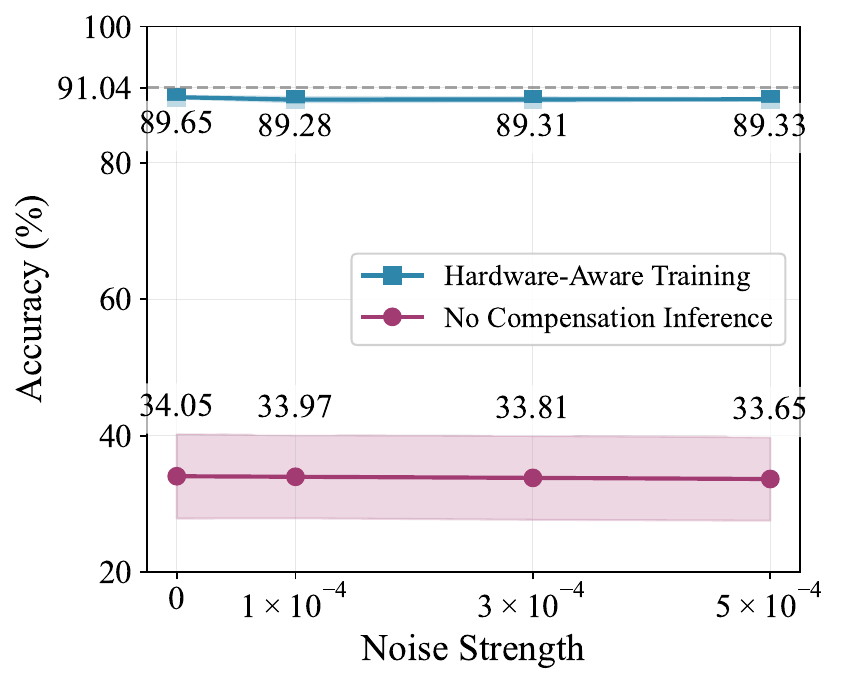}
\caption{Accuracy}
\label{fig:drift_alpha_accuracy}
\end{subfigure}
\hfill
\begin{subfigure}[b]{0.32\textwidth}
\centering
\includegraphics[width=\linewidth]{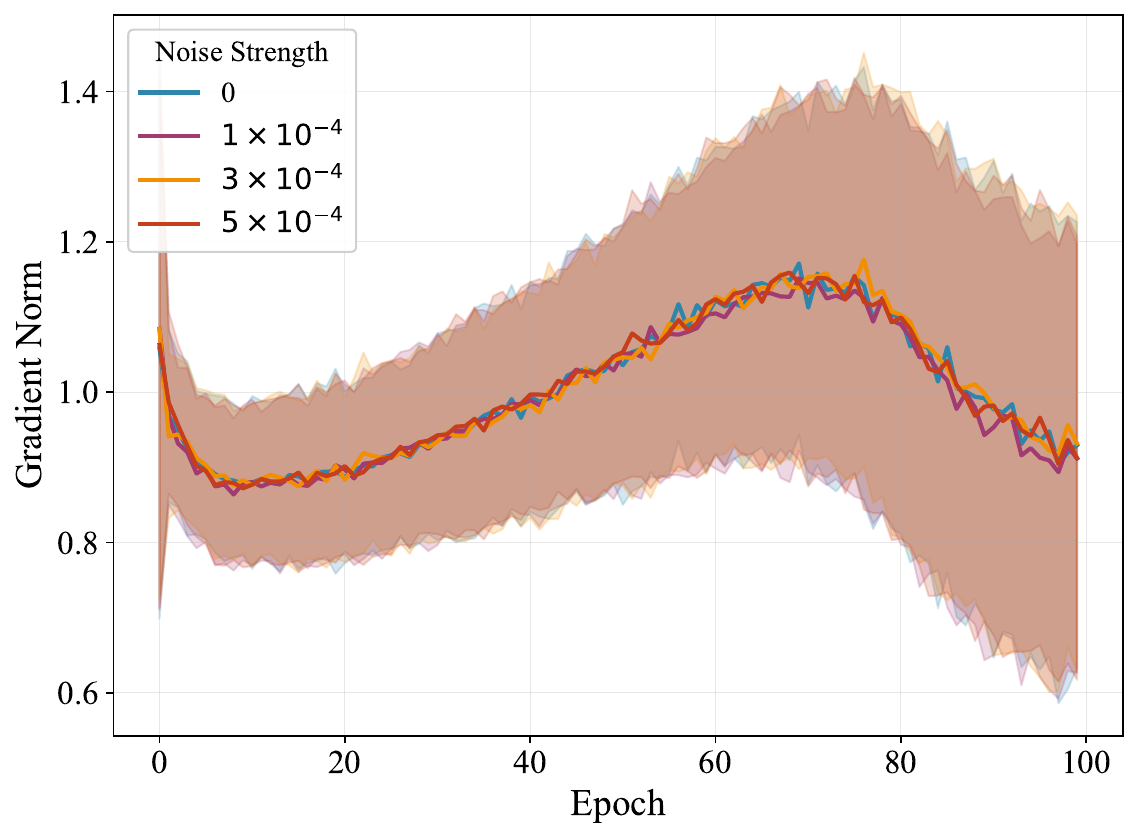}
\caption{Gradient Norm}
\label{fig:drift_alpha_grad}
\end{subfigure}
\hfill
\begin{subfigure}[b]{0.32\textwidth}
\centering
\includegraphics[width=\linewidth]{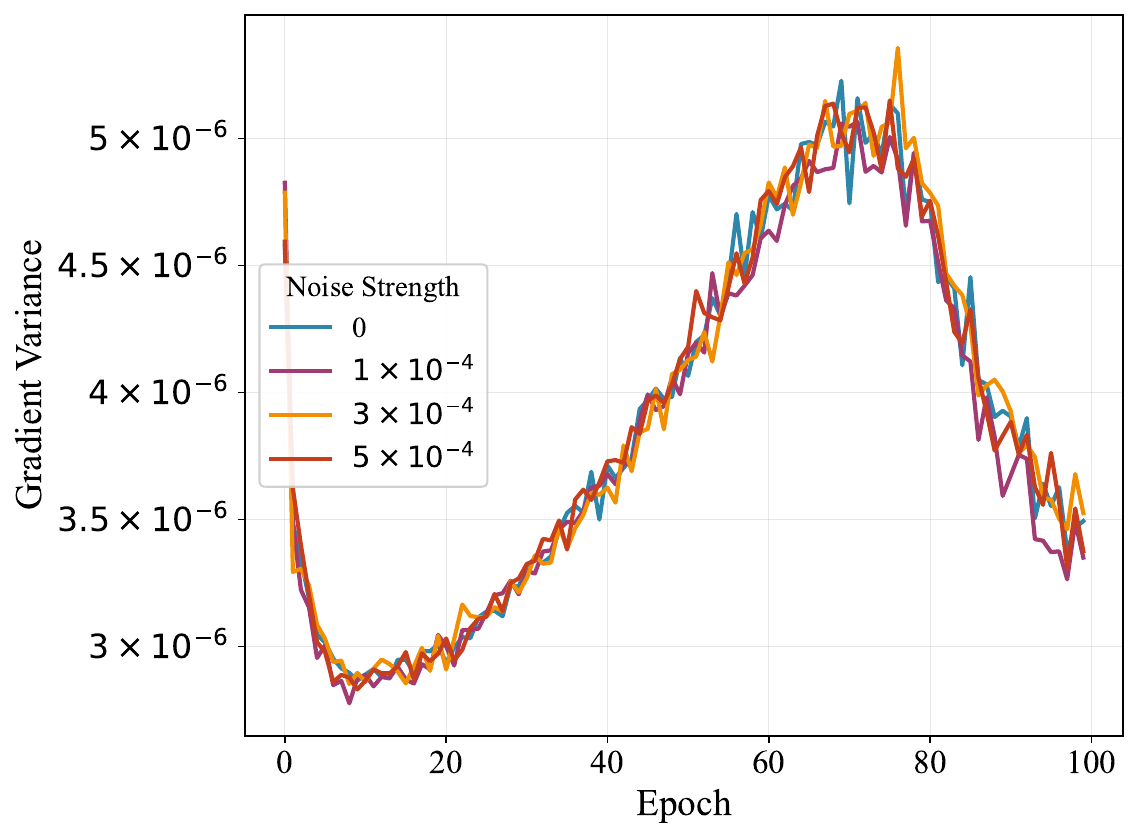}
\caption{Gradient Variance}
\label{fig:drift_alpha_grad_var}
\end{subfigure}
\caption{Accuracy, gradient norm and gradient variance under deterministic conductance drift for different attenuation rates $\alpha$.}
\label{fig:drift_alpha}
\end{figure*}

\subsubsection{ADC Discretization With STE}
\label{sec:ADC_with_STE}

\begin{figure*}[h]
\centering
\begin{subfigure}[b]{0.4\textwidth}
\centering
\includegraphics[width=\linewidth]{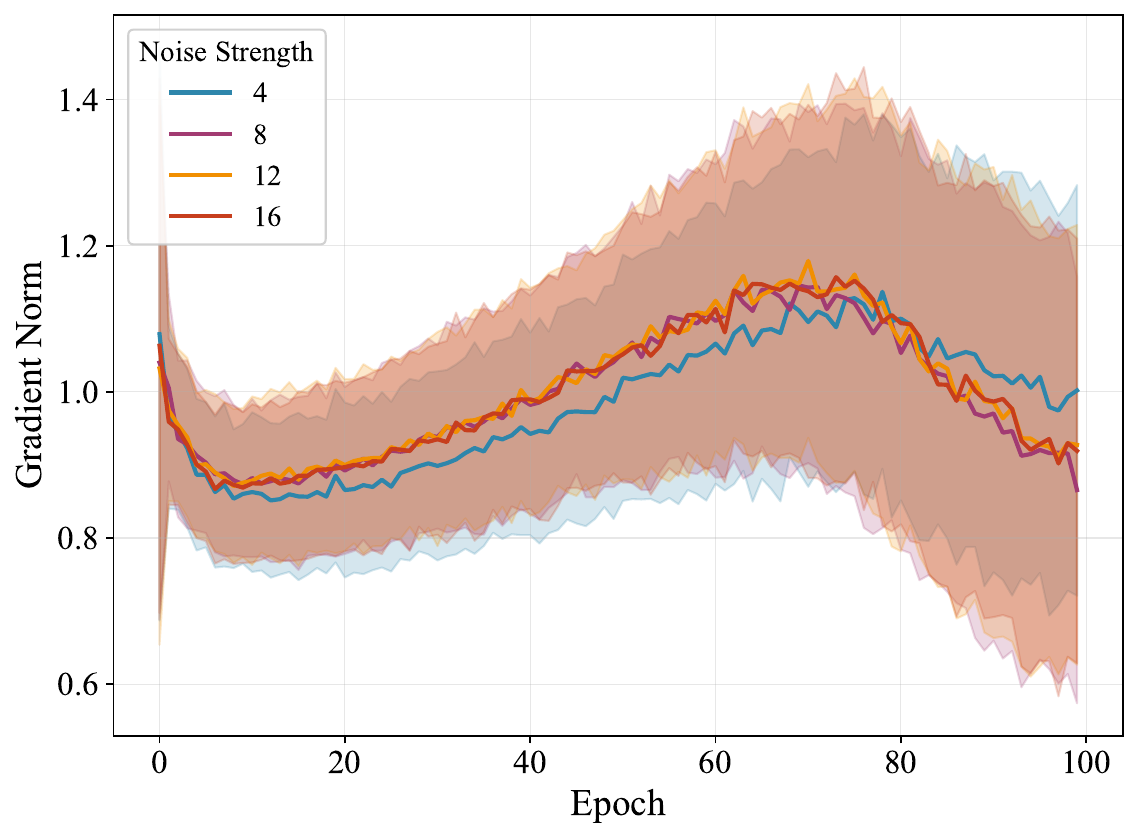}
\caption{Gradient Norm}
\label{fig:ADC_with_STE_grad}
\end{subfigure}
\hfill
\begin{subfigure}[b]{0.4\textwidth}
\centering
\includegraphics[width=\linewidth]{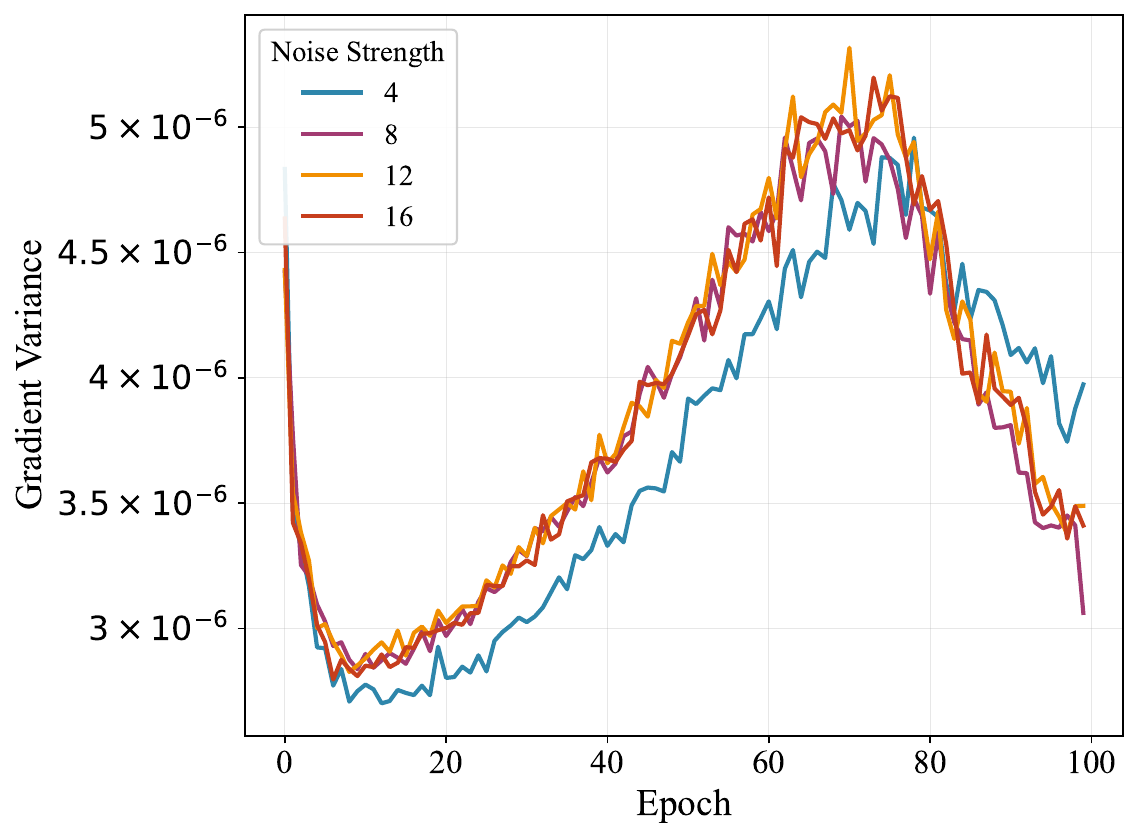}
\caption{Gradient Variance}
\label{fig:ADC_with_STE_grad_var}
\end{subfigure}
\caption{The restoration of gradient flow via STE demonstrates that gradient accessibility, not bit-width, dictates learnability.}
\label{fig:ADC_with_STE}
\end{figure*}

\subsubsection{Results on CIFAR-100}

\begin{figure*}[htb]
\centering
\begin{subfigure}[b]{0.32\textwidth}
\centering
\includegraphics[width=\linewidth]{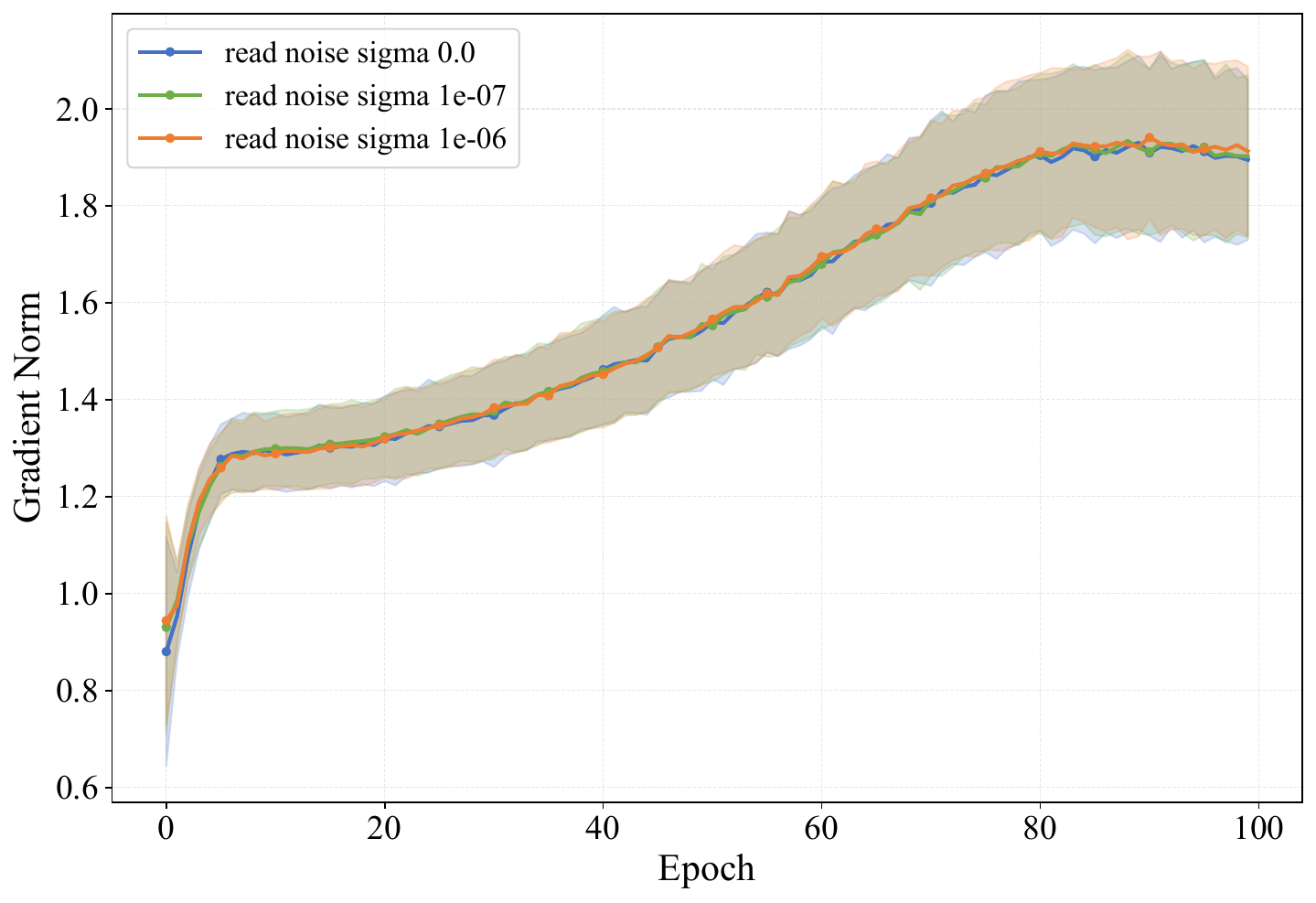}
\caption{Additive: $\sigma_r$}
\label{fig:grad-additive-cifar100}
\end{subfigure}
\hfill
\begin{subfigure}[b]{0.32\textwidth}
\centering
\includegraphics[width=\linewidth]{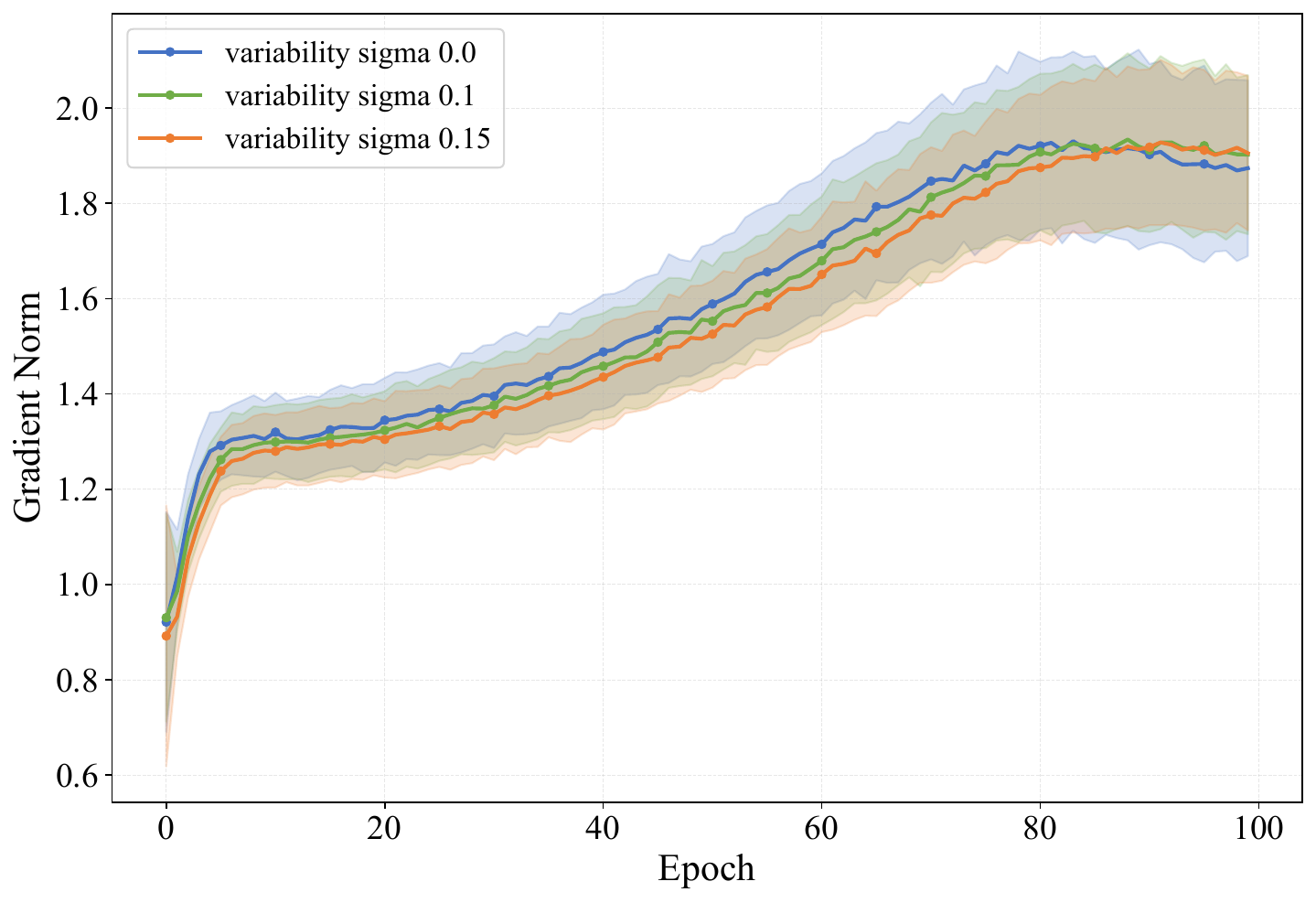}
\caption{Multiplicative: $\sigma_v$}
\label{fig:grad-multiplicative-cifar100}
\end{subfigure}
\hfill
\begin{subfigure}[b]{0.32\textwidth}
\centering
\includegraphics[width=\linewidth]{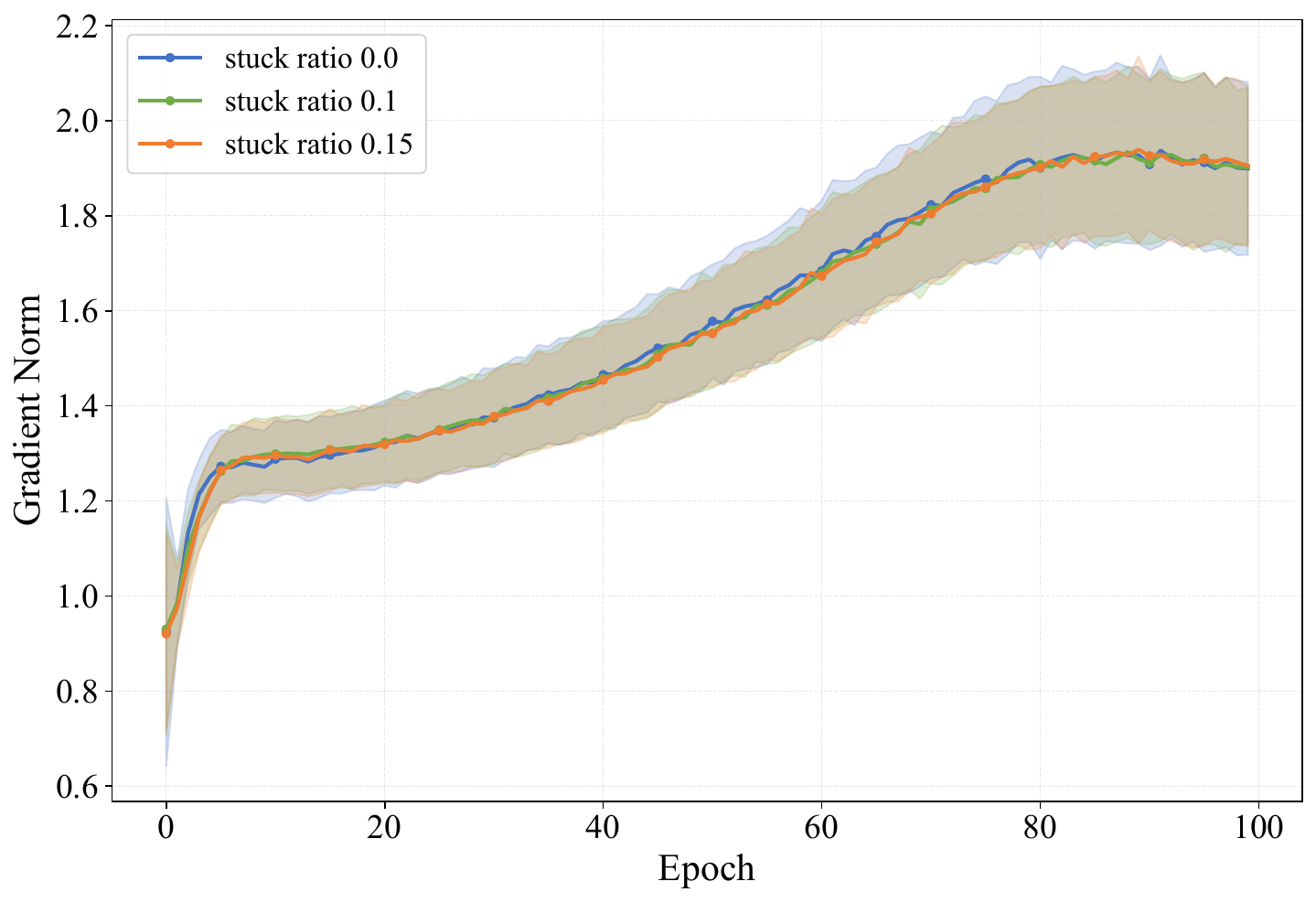}
\caption{Projection: $\rho$}
\label{fig:grad-projection-cifar100}
\end{subfigure}
\hfill
\begin{subfigure}[b]{0.32\textwidth}
\centering
\includegraphics[width=\linewidth]{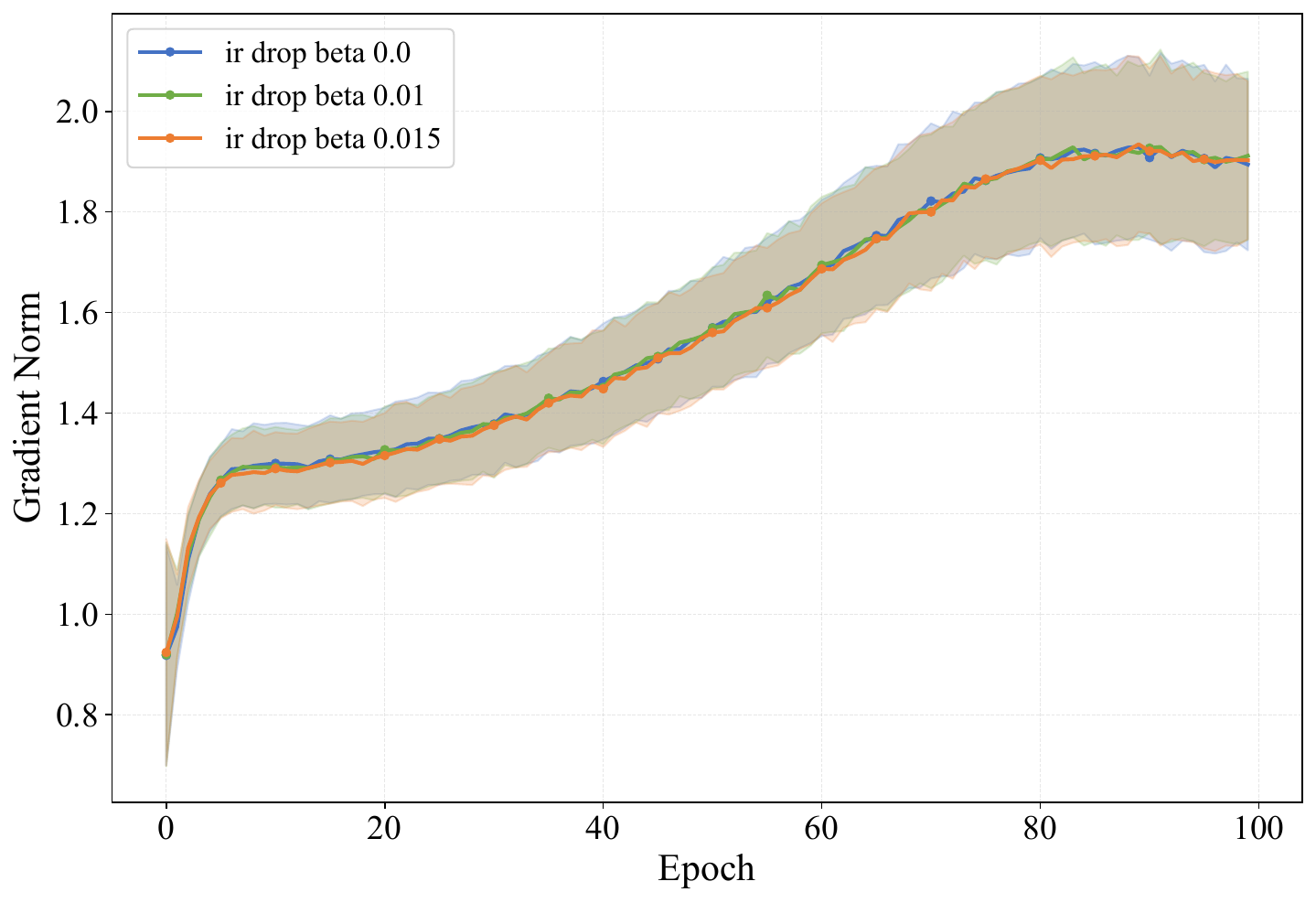}
\caption{Input scaling: $\beta$}
\label{fig:grad-input-scaling-cifar100}
\end{subfigure}
\hfill
\begin{subfigure}{0.32\textwidth}
\centering
\includegraphics[width=\linewidth]{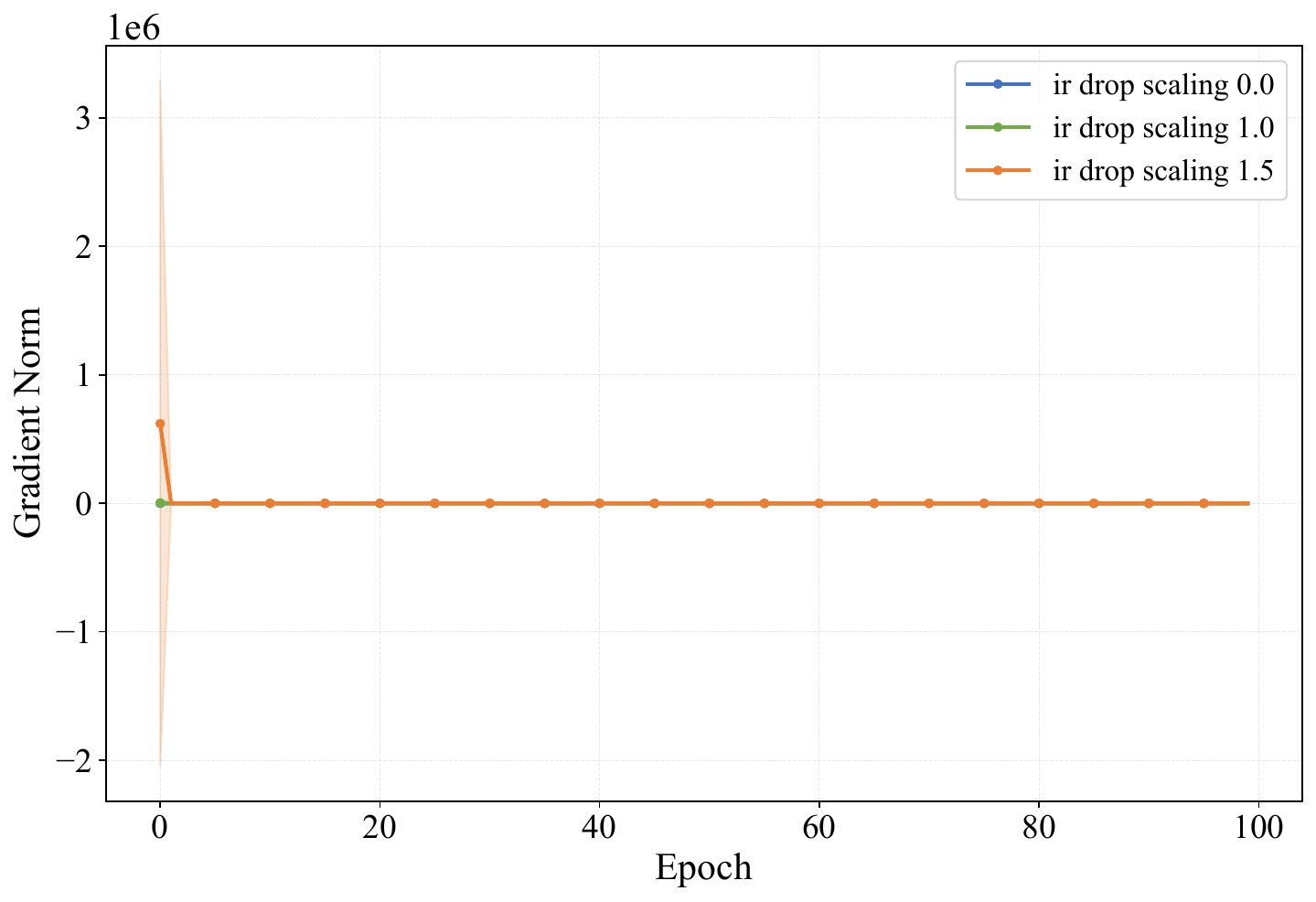}
\caption{Strongly coupled}
\label{fig:grad-strongly-coupled-cifar100}
\end{subfigure}
\hfill
\begin{subfigure}{0.32\textwidth}
\centering
\includegraphics[width=\linewidth]{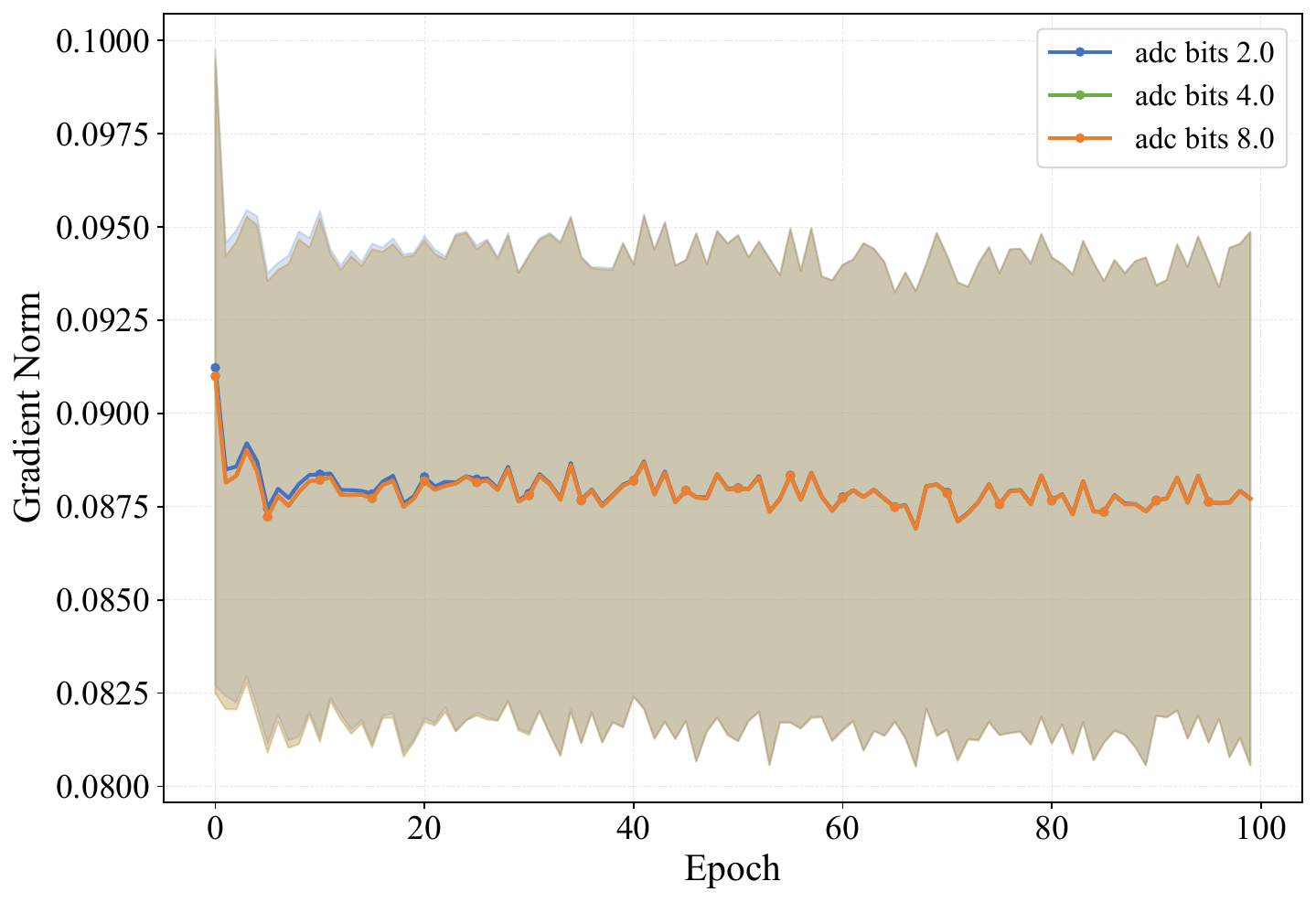}
\caption{Discretization}
\label{fig:grad-discretization-cifar100}
\end{subfigure}
\caption{
\textbf{Gradient norm dynamics across perturbation classes with varying strengths on CIFAR-100.}
(a) Additive perturbations (read noise $\sigma_r$). (b) Multiplicative perturbations (variability $\sigma_v$). (c) Projection perturbations (stuck-at ratio $\rho$). (d) Input-dependent scaling (IR-drop $\beta$). (e) Strongly coupled nonlinear IR-drop. (f) Discretization via ADC quantization. Each subplot shows gradient norm (solid line) and standard deviation (shaded region) across four perturbation strengths.}
\label{fig:gradient-dynamics-cifar100}
\end{figure*}

\begin{figure*}[htb]
\centering
\begin{subfigure}[b]{0.32\textwidth}
\centering
\includegraphics[width=\linewidth]{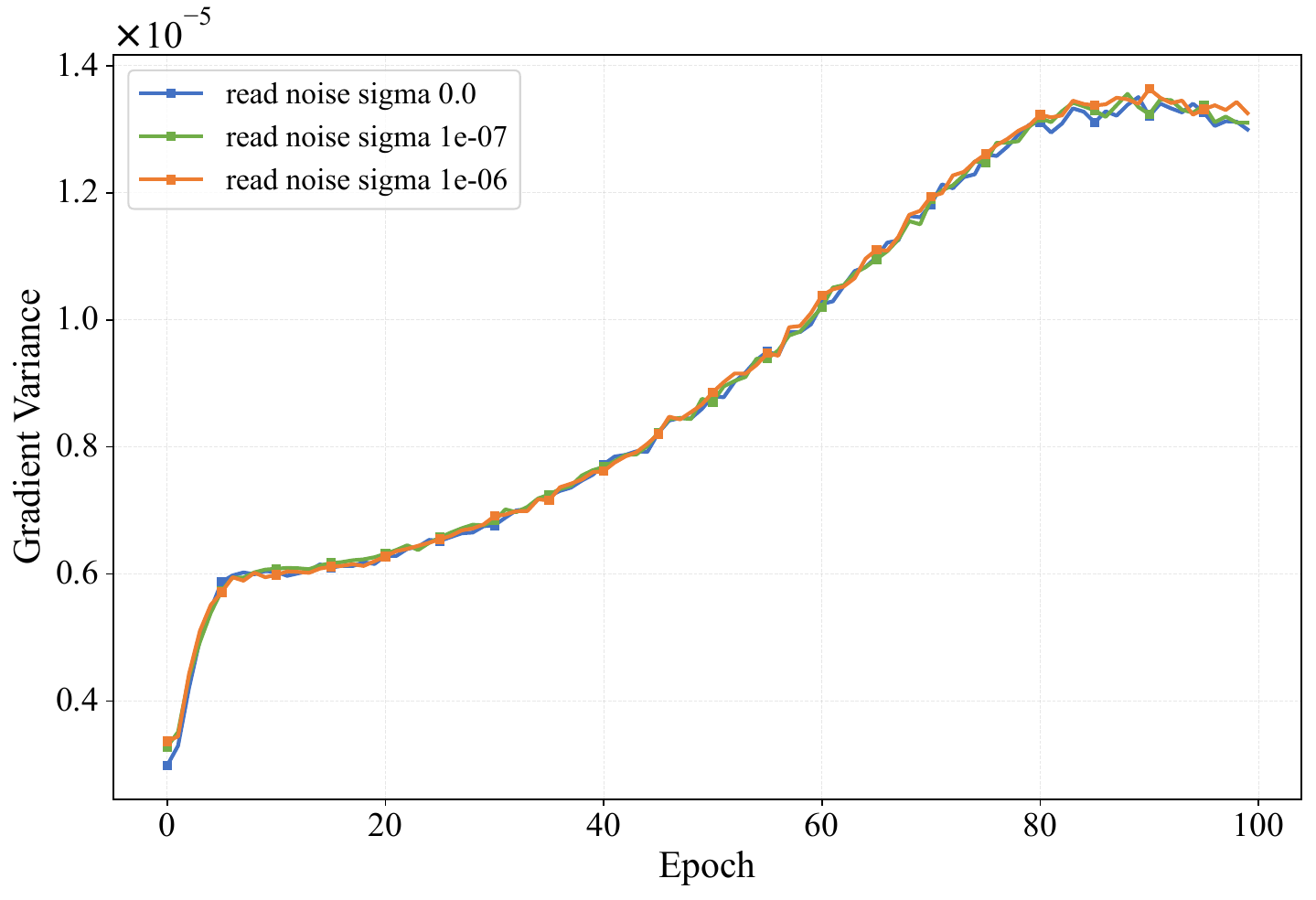}
\caption{Additive: $\sigma_r$}
\label{fig:grad-var-additive-cifar100}
\end{subfigure}
\hfill
\begin{subfigure}[b]{0.32\textwidth}
\centering
\includegraphics[width=\linewidth]{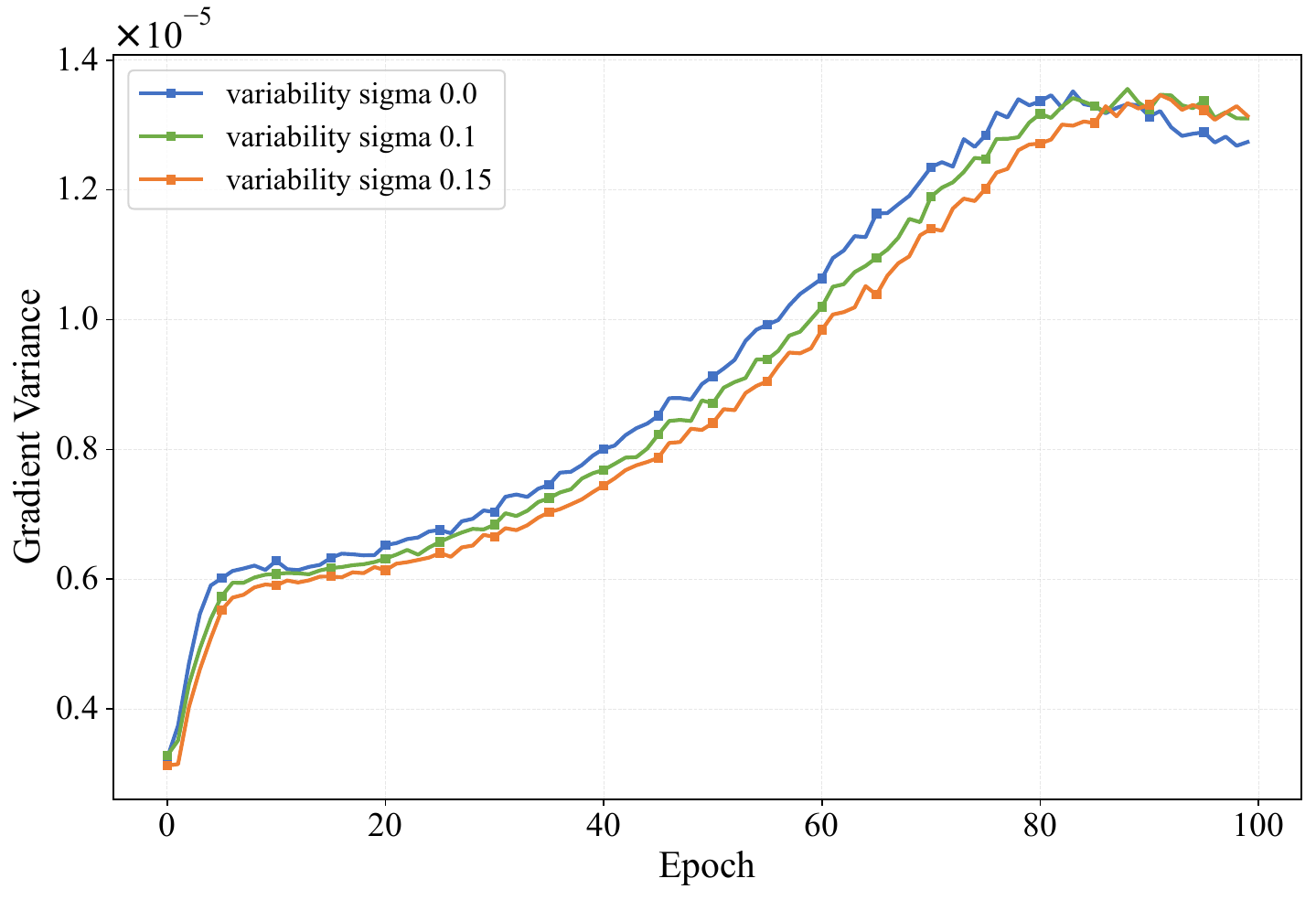}
\caption{Multiplicative: $\sigma_v$}
\label{fig:grad-var-multiplicative-cifar100}
\end{subfigure}
\hfill
\begin{subfigure}[b]{0.32\textwidth}
\centering
\includegraphics[width=\linewidth]{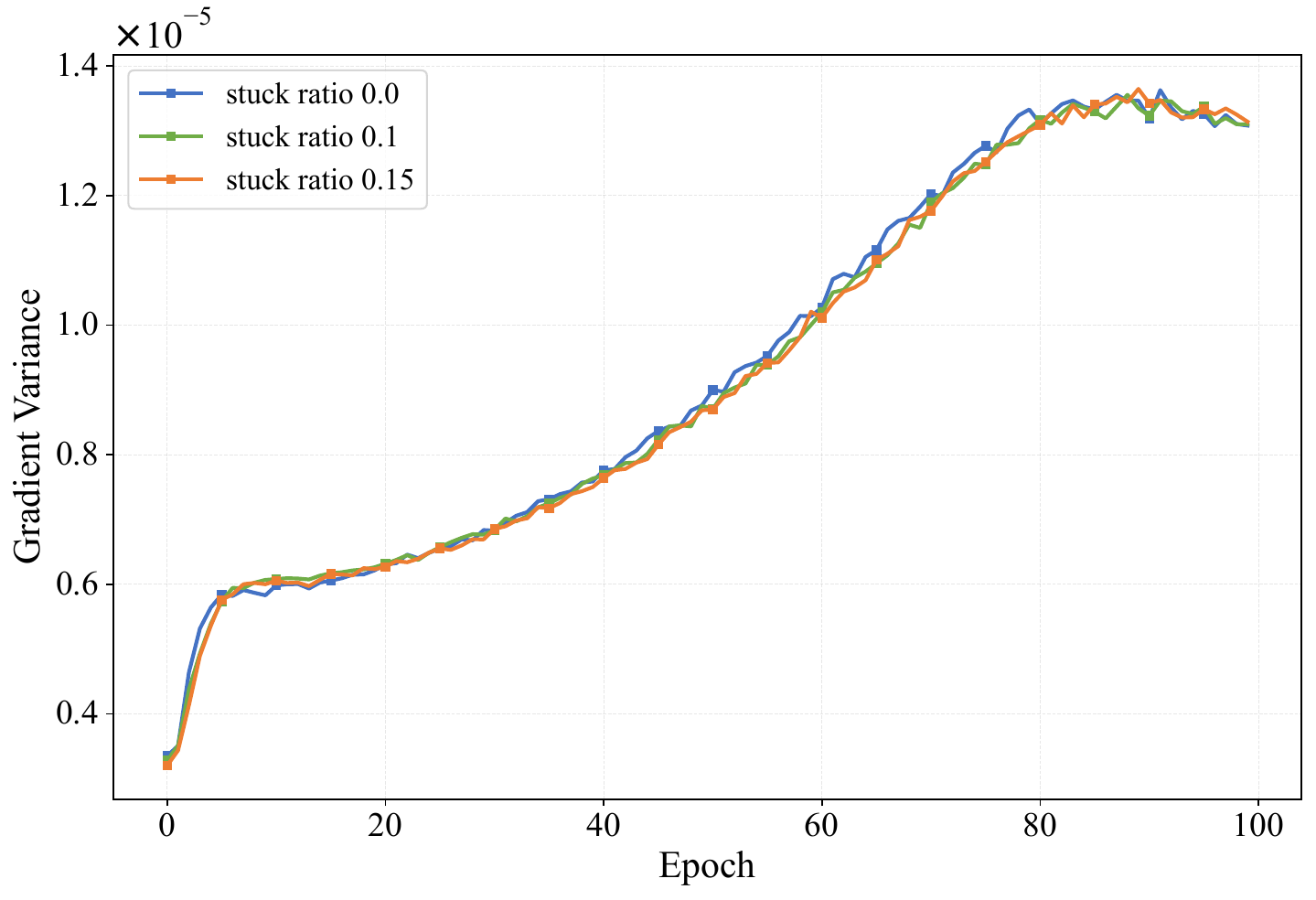}
\caption{Projection: $\rho$}
\label{fig:grad-var-projection-cifar100}
\end{subfigure}
\hfill
\begin{subfigure}[b]{0.32\textwidth}
\centering
\includegraphics[width=\linewidth]{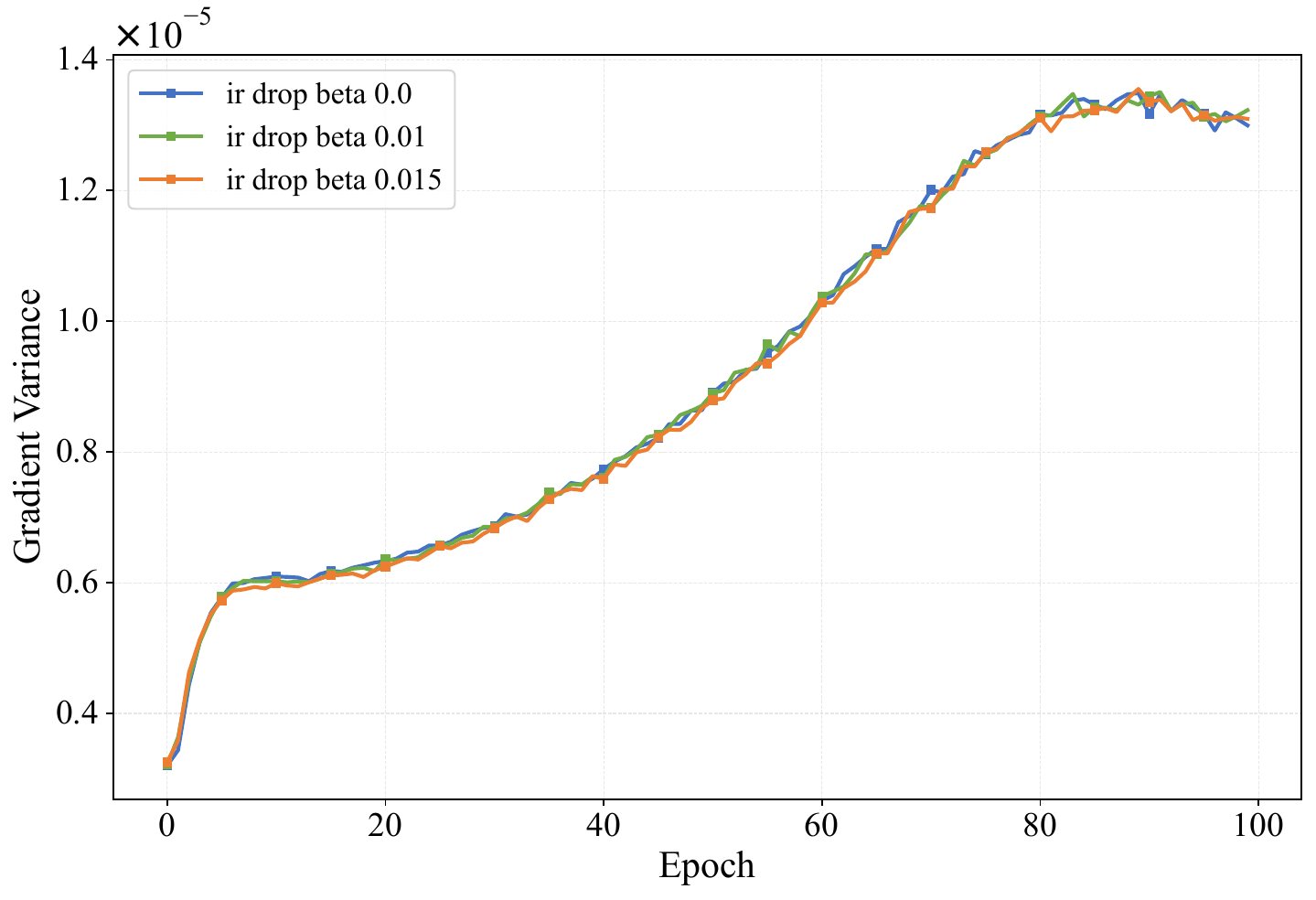}
\caption{Input scaling: $\beta$}
\label{fig:grad-var-input-scaling-cifar100}
\end{subfigure}
\hfill
\begin{subfigure}{0.32\textwidth}
\centering
\includegraphics[width=\linewidth]{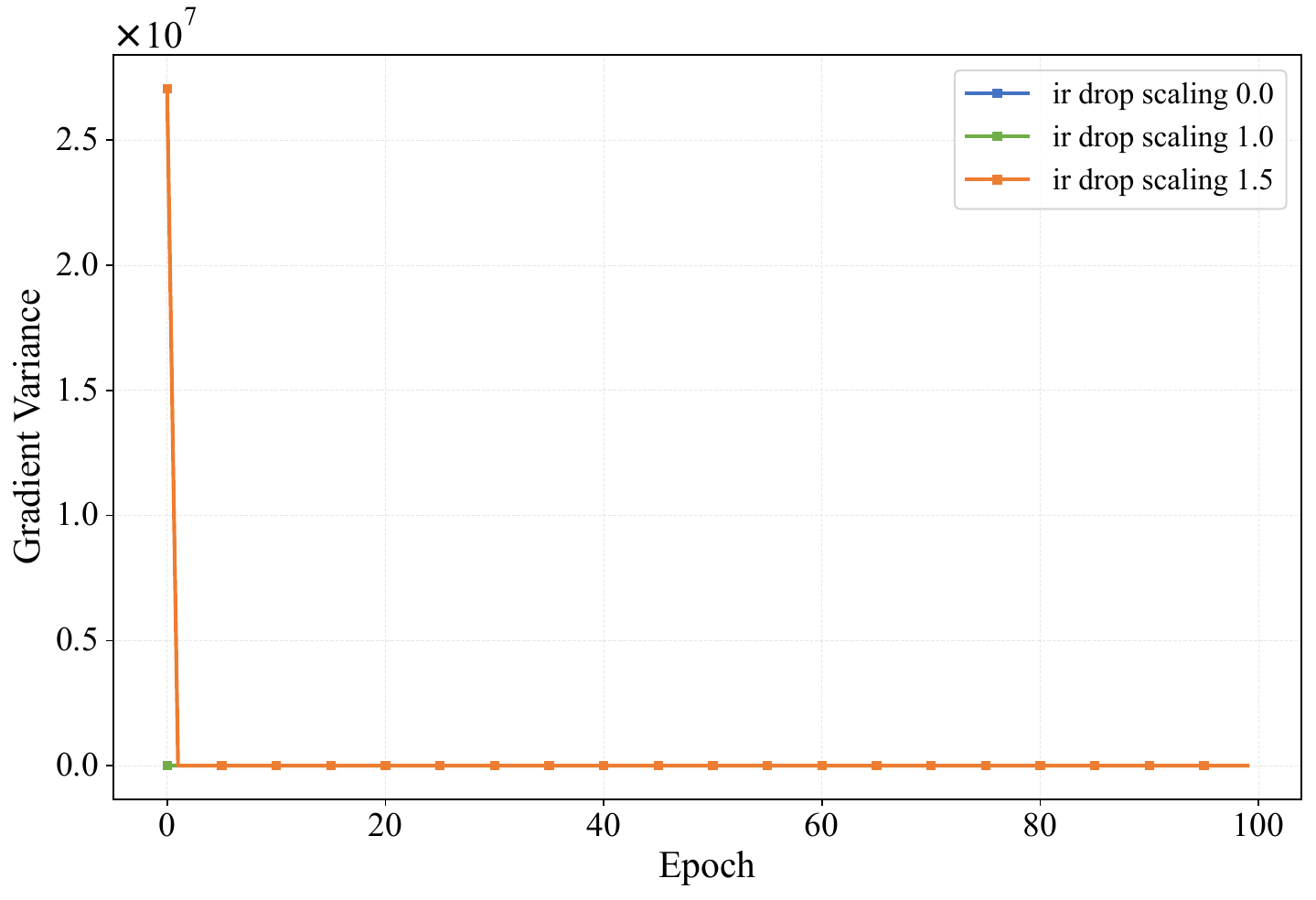}
\caption{Strongly coupled}
\label{fig:grad-var-strongly-coupled-cifar100}
\end{subfigure}
\hfill
\begin{subfigure}{0.32\textwidth}
\centering
\includegraphics[width=\linewidth]{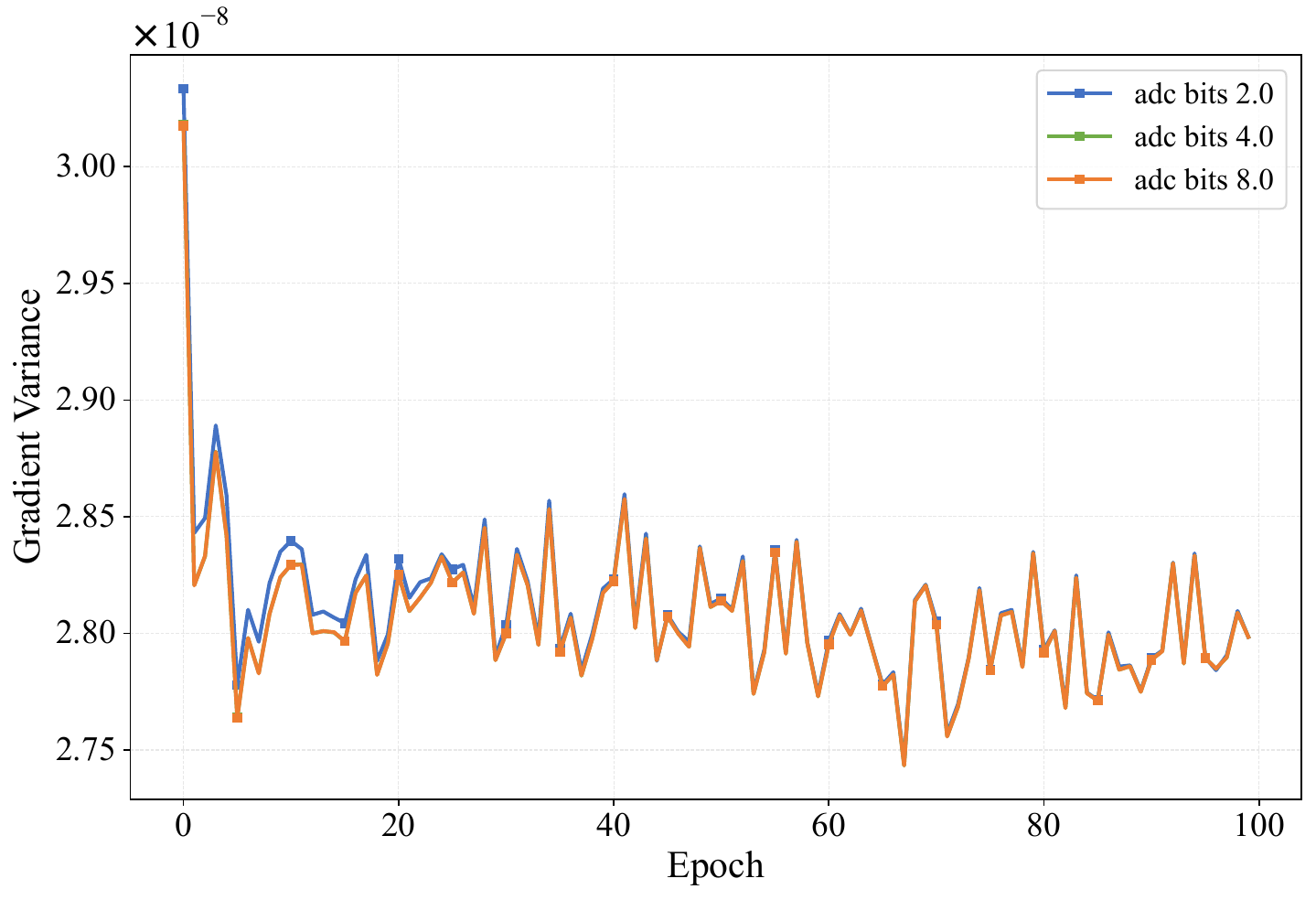}
\caption{Discretization}
\label{fig:grad-var-discretization-cifar100}
\end{subfigure}
\caption{
\textbf{Gradient variance across perturbation classes with varying strengths on CIFAR-100.}
(a) Additive perturbations (read noise $\sigma_r$). (b) Multiplicative perturbations (variability $\sigma_v$). (c) Projection perturbations (stuck-at ratio $\rho$). (d) Input-dependent scaling (IR-drop $\beta$). (e) Strongly coupled nonlinear IR-drop. (f) Discretization via ADC quantization.}
\label{fig:gradient-variance-cifar100}
\end{figure*}

\end{document}